\UseRawInputEncoding
\documentclass[12pt,3p]{elsarticle}
\usepackage{booktabs}
\usepackage{makecell}
\usepackage[utf8]{inputenc}
\usepackage{url}
\usepackage{graphicx}
\usepackage{balance}
\usepackage{amsmath}
\usepackage{subfigure}
\usepackage{graphicx}
\usepackage{balance}
\usepackage{color}
\usepackage{comment}
\usepackage{array}
\usepackage{lipsum}
\usepackage{caption}
\usepackage{hyperref}
 \usepackage{blindtext}
\usepackage{enumitem}
\usepackage{lipsum}
\usepackage{capt-of}
\usepackage{tikz}
\usetikzlibrary{arrows,positioning,automata}
\usepackage{balance}
\usepackage{amsmath}
\usepackage[T1]{fontenc}
\usepackage{algorithm}
\usepackage{algpseudocode}
\usepackage{subfigure}
\usepackage{graphicx}
\usepackage[utf8]{inputenc}
\usepackage{balance}
\usepackage{multirow}
\usepackage{multicol}
\usepackage{color}
\usepackage{comment}
\usepackage{array}
\usepackage{soul}
\usepackage[justification=centering]{caption}
\usepackage[lighttt]{lmodern}
\usepackage{libertine}
\usepackage{amssymb}
\usepackage{float}
\usepackage{hyperref}
\hypersetup{
    colorlinks=true,
    linkcolor=black,
    urlcolor=blue,
    citecolor=black
}
\journal{Elsevier}

\begin{document}
    
    
    

\begin{frontmatter}
\title{Large Language Models Meet Stance Detection: A Survey of Tasks, Methods, Applications, Challenges, and Future Directions 
 }

\address[1]{Department of Computer Science and Engineering, Indian Institute of Technology (IIT) Indore, Indore 453552, India}

\author[1]{Lata Pangtey\textsuperscript{†,}}
\ead{ms2304101009@iiti.ac.in}

\author[1]{Anukriti Bhatnagar\textsuperscript{†,}}
\ead{mt2302101007@iiti.ac.in}
\author[1]{Shubhi Bansal}
\ead{phd2001201007@iiti.ac.in}
\author[1]{Shahid Shafi Dar}
\ead{phd2201201004@iiti.ac.in}
\author[1]{Nagendra Kumar\corref{cor1}}
\cortext[cor1]{Corresponding author}
\ead{nagendra@iiti.ac.in}
\def\thefootnote{†}\footnotetext{These authors contributed equally to this work}

\begin{abstract}
Stance detection determines a user's opinion toward a particular target or statement. The task is essential for understanding subjective content across various platforms such as social media, news articles, and online reviews.
The rapid growth of user-generated content has made stance detection a critical task in natural language processing.
Recent advances in large language models have revolutionized stance detection by introducing novel capabilities in contextual understanding, cross-domain generalization, and multimodal analysis. Despite these progressions, existing surveys often lack comprehensive coverage of approaches that specifically leverage large language models (LLMs) for stance detection. To bridge this critical gap, our review article conducts a systematic analysis of stance detection, comprehensively examining recent advancements of LLMs transforming the field, including foundational concepts, methodologies, datasets, applications, and emerging challenges.
We present a novel taxonomy for LLM-based stance detection approaches, structured along three key dimensions: 1) learning methods, including supervised, unsupervised, few-shot, and zero-shot; 2) data modalities, such as unimodal, multimodal, and hybrid; and 3) target relationships, encompassing in-target, cross-target, and multi-target scenarios.
Furthermore, we discuss the evaluation techniques and
analyze benchmark datasets and performance trends, highlighting the strengths and limitations of different architectures. Key applications in misinformation detection, political analysis, public health monitoring, and social media moderation are discussed. Finally, we identify critical challenges such as implicit stance expression, cultural biases, and computational constraints, while outlining promising future directions, including explainable stance reasoning, low-resource adaptation, and real-time deployment frameworks. Our survey highlights emerging trends, open challenges, and future directions to guide researchers and practitioners in developing next-generation stance detection systems powered by large language models.

\end{abstract}

\end{frontmatter}
\newpage
\tableofcontents
\newpage
\section{Introduction}
The rapid expansion of social media and online communication has led to generation of huge amount of opinion-based content.
Stance detection, the computational process of identifying an author's viewpoint toward a specific target, such as a policy, product, or ideology, has become indispensable for applications ranging from political discourse analysis to public health monitoring.
Stance detection is the computational task of determining the position or attitude expressed in text toward a specific target, which could be a person, policy, product, or idea.
As illustrated in Table \ref{table:example}, stance detection categorizes textual expressions into labels such as favor, against and none across diverse contexts.
The emergence of large language models (LLMs) has revolutionized the task of stance detection by offering new capabilities in understanding complex opinions, cross-domain generalization, and multimodal stance analysis. This survey provides a comprehensive examination of how LLMs are transforming stance detection, addressing both the technological advancements and the practical challenges that remain.
Stance detection requires understanding the relationship between a text and a particular target.
The advent of transformer-based large language models like BERT, GPT, and their successors has marked a paradigm shift in NLP capabilities.

\begin{figure*}[!ht]
    \centering
    \captionsetup{justification=justified}   \includegraphics[scale=0.5]{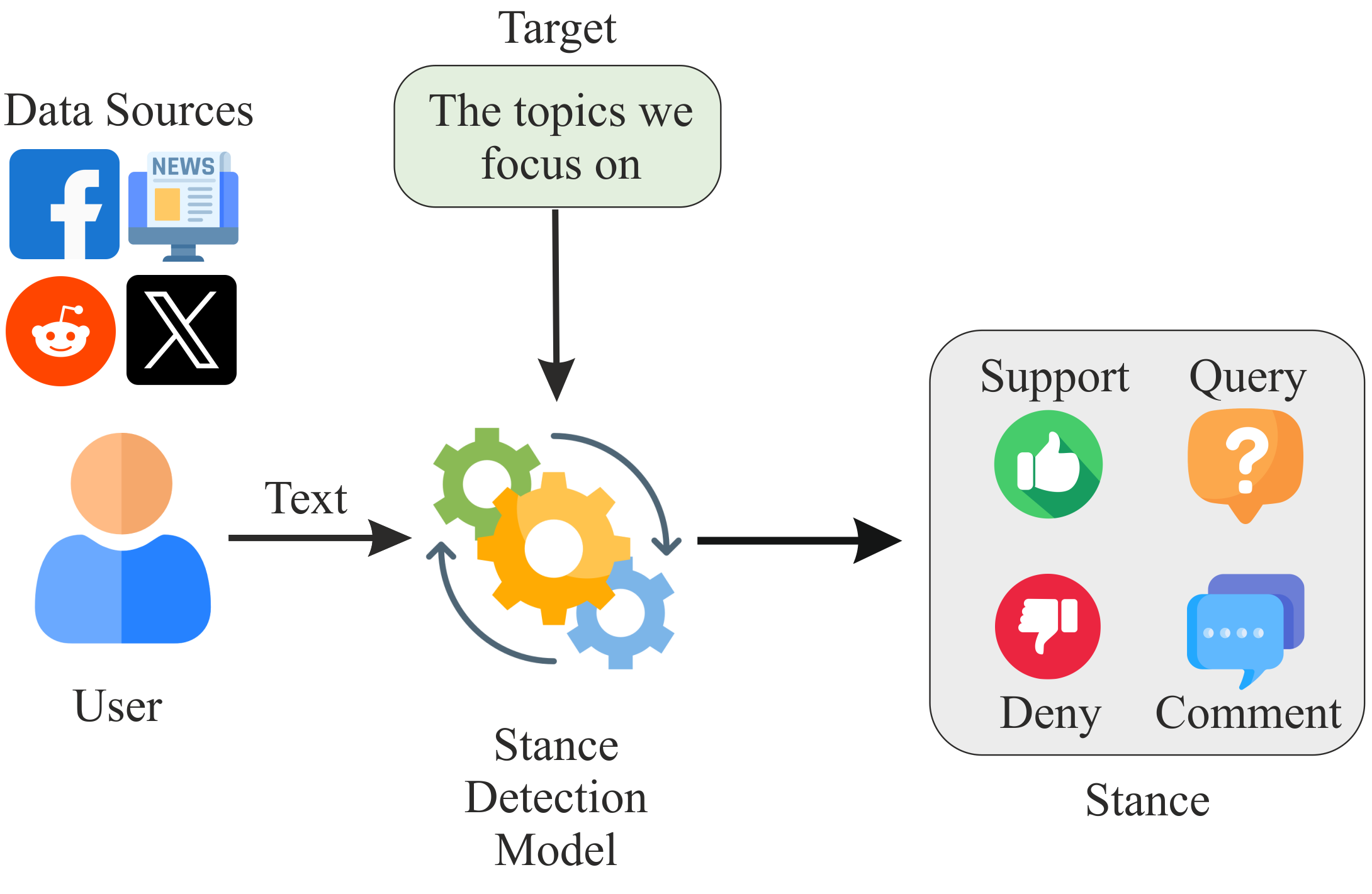}
  
    \caption{ General Framework for Stance Detection}
    \label{fig:fig1}
\end{figure*}

\begin{figure*}[!ht]
    \centering
    \captionsetup{justification=justified}   \includegraphics[width=15cm, height=8.5cm]{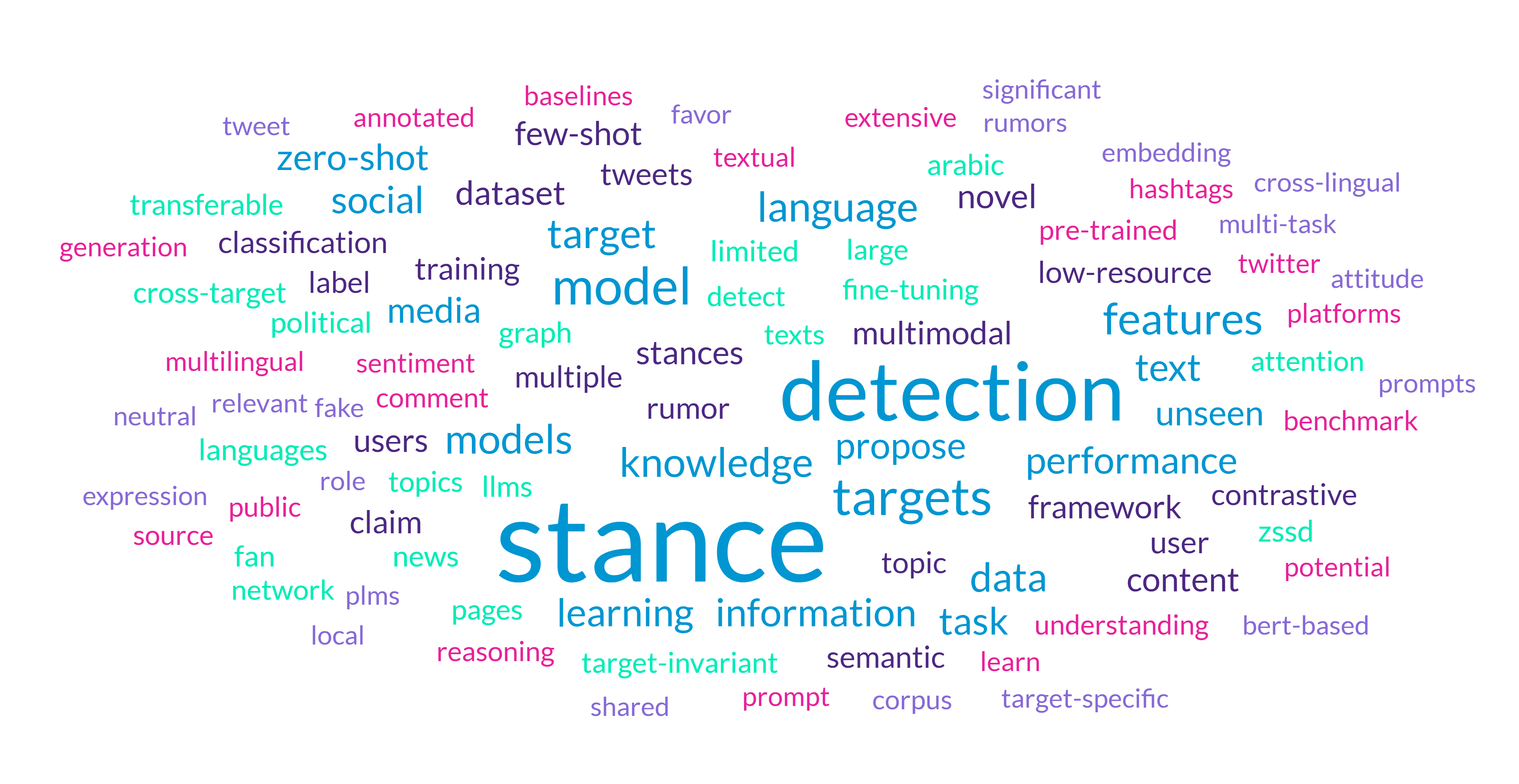}
    \caption{Term Frequency Word Cloud Summarizing Prominent Vocabulary in Stance Detection Research}
    \label{fig:fig2}
\end{figure*}

\begin{table}[!ht]\normalsize
    \centering
    \caption{Examples of Target-Text and the Stance}
    \begin{tabular}{|p{6.7cm}|p{6.2cm}|p{1.6cm}|}
    \hline
   \textbf{Target} & \textbf{Text} & \textbf{Stance} \\ \hline
    Legalization of Abortion & Life is sacred on all levels. Abortion does not compute with my philosophy (Red on \#OITNB). & Against \\ \hline
    Climate Change is the real concern& The biggest terror threat in the World is climate change \#drought \#floods. & Favor \\ \hline
    Feminist Movement & Gender research helps to develop nursing research theoretical frameworks, concepts and methodology 
 \@genderanded 
 \#research \#SemST. & None \\ \hline
    \end{tabular}
    \label{table:example}
\end{table}
Stance detection is a fundamental task in natural language processing (NLP) that aims to determine the position or attitude expressed in a text towards a particular target, claim, or topic. Unlike sentiment analysis, which focuses on general emotional tone, stance detection deals with whether the author supports, opposes, queries, or remains neutral about a specific subject. This task has gained prominence in the context of social media analysis, misinformation tracking, political discourse, and public opinion mining, where identifying implicit or explicit stances is critical for downstream applications such as rumor verification and argument mining.  Traditional approaches relied on handcrafted features, syntactic patterns, and supervised machine learning models, but these often struggled with generalization across domains and ambiguity in short, informal texts.  As depicted in Figure~\ref{fig:fig1}, the general framework for Stance Detection provides a comprehensive approach to understanding and identifying stances in texts.
Recent advances in large language models (LLMs) have significantly reshaped stance detection methodologies. Pretrained transformers such as BERT, RoBERTa, and domain-specific variants like StanceBERTa have demonstrated superior capability in modelling contextual dependencies and semantic nuances of texts. 
A visual summary of frequently occurring stance-related terms is presented in Figure~\ref{fig:fig2}, illustrating common lexical patterns associated with various stances.
LLM models can implicitly capture relationships between the input statement and the target stance, even in the absence of explicit indicators, by leveraging transfer learning from vast corpora \cite{akash-etal-2025-large, 10.1145/3716856}. LLM-based methods also enable more flexible adaptation to multilingual settings, domain shifts, and few-shot learning scenarios, making them a powerful foundation for building robust and scalable stance detection systems. 
This shift has opened new frontiers for research, particularly in incorporating commonsense reasoning, emotion-aware representations, and multitask learning into stance prediction pipelines.
The LLM models are pre-trained on massive text corpora using self-supervised objectives, demonstrate remarkable abilities in understanding context, capturing subtle linguistic patterns, and performing few-shot learning. 
While several surveys \cite{10.1145/3369026,ALDAYEL2021102597,10.1145/3488560.3501391, zhang2024survey, 10.1145/3744641} have examined stance detection and the applications of LLMs in NLP, our work makes distinct contributions:

\begin{itemize}
\item \textbf{LLM-Centric Focus}: Unlike traditional surveys that broadly cover stance detection, our work specifically analyzes how large language models (LLMs) have transformed each component of the stance detection pipeline—from feature extraction to cross-target generalization. To highlight this evolution, we compile and examine the number of research papers published between 2019 and April 2025 that employ LLM-based techniques for stance detection. As illustrated in Figure~\ref{fig:fig3}, the number of such studies has increased significantly over this period. Figure~\ref{fig:fig4} further shows a pie chart of the distribution of these collected papers across different publication venues and repositories, highlighting the diversity of sources we considered.
This includes models such as BERT, RoBERTa, and more recent generative models like GPT. Our analysis reveals a clear upward trend, reflecting the growing reliance on LLMs in addressing stance detection challenges.

\begin{figure}[htbp]
  \centering
  \begin{minipage}[b]{0.5\textwidth}
    \includegraphics[width=\textwidth]{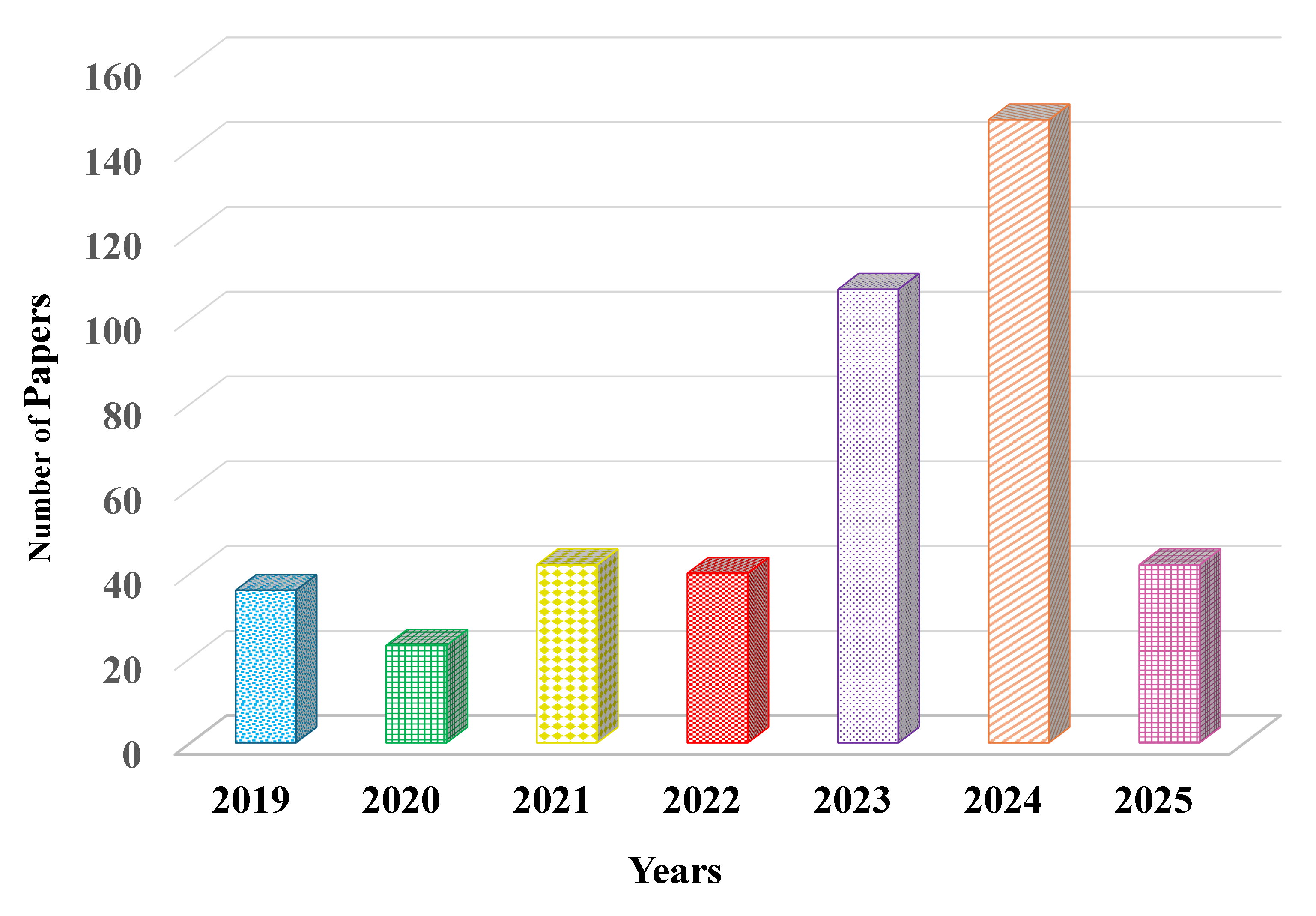}
    \caption{Number of LLM-based Stance Detection Studies between 2019 and 2025}
    \label{fig:fig3}
  \end{minipage}
  \hfill
  \begin{minipage}[b]{0.45\textwidth}
    \includegraphics[width=\textwidth]{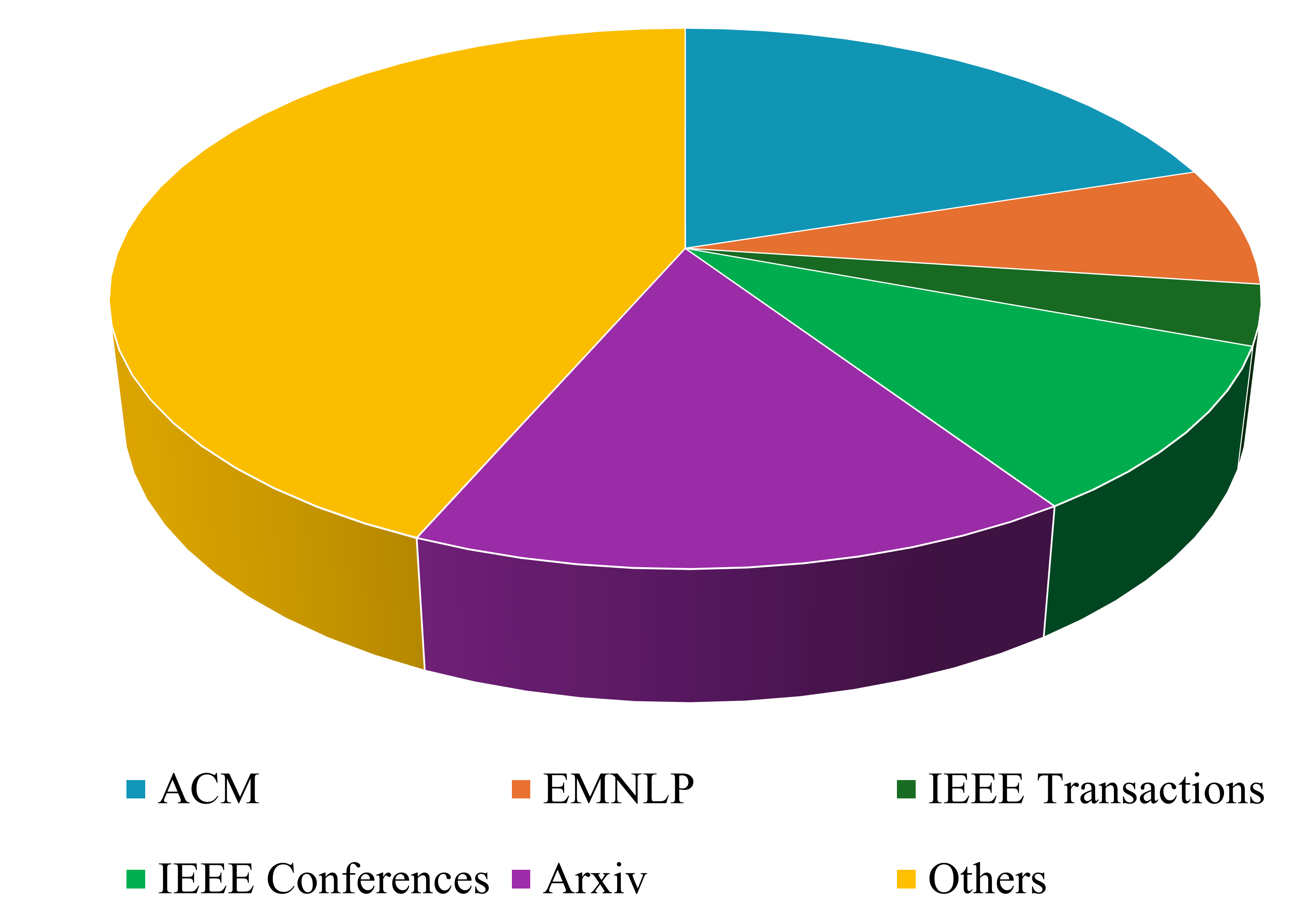}
    \caption{Venue Distribution of Publications Related to Stance Detection
}
    \label{fig:fig4}
  \end{minipage}
\end{figure}
\item \textbf{Comprehensive Taxonomy of LLM-Based Stance Detection Approaches}: We introduce a novel taxonomy that categorizes LLM-based stance detection approaches along three key aspects. First, learning methods, which include supervised, few-shot, and zero-shot learning paradigms. Second, data modalities, distinguishing between unimodal and multimodal inputs. Third, target relationships, which capture the stance detection scope across in-target, cross-target, and multi-target scenarios. 

\item \textbf{Analysis of Benchmarks, Performance Trends, and Practical Challenges}:
The paper provides a critical analysis of benchmarks, performance trends, and practical challenges in stance detection. It curates and compares over 40 datasets spanning multiple languages and modalities. Performance benchmarking reveals F1-macro scores of several stance detection datasets. The paper also identifies limitations such as implicit stance expression, cultural biases in training data, and the computational costs of LLMs. This evaluation helps researchers select models suited for specific use cases and highlights underexplored areas like low-resource language adaptation.
\end{itemize}
In the subsequent sections, we present a detailed analysis of seven years of research in stance detection, evaluated across LLM techniques. 
The overall structure and flow of the paper are illustrated in Figure~\ref{fig:fig5}, which serves as a visual outline of the key components covered.
\begin{figure*}[!ht]
    \centering
    \captionsetup{justification=justified}   \includegraphics[scale=0.61]{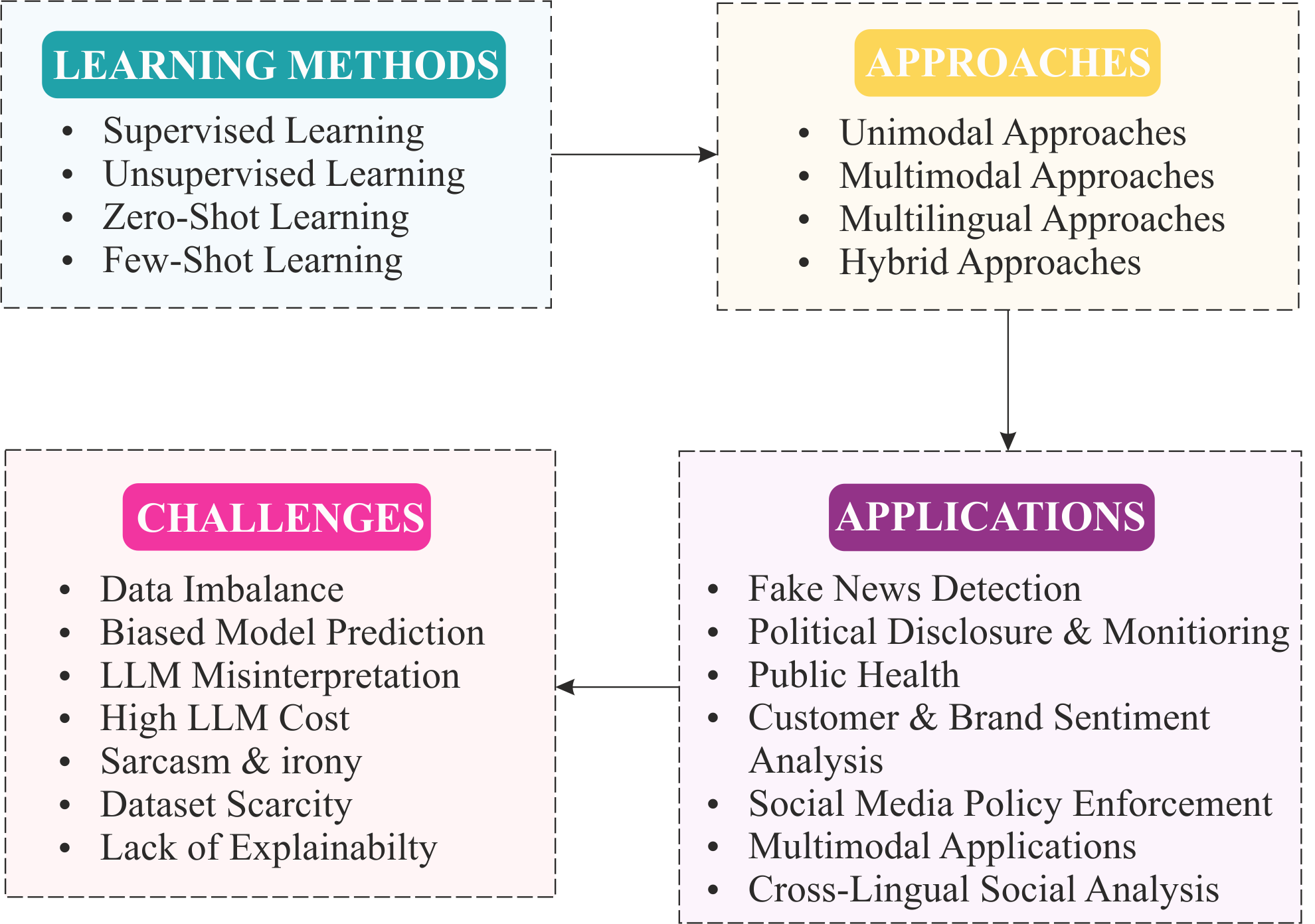}
  
    \caption{ Outline of the Survey on Stance Detection}
    \label{fig:fig5}
\end{figure*}

Section \ref{Section:llm} provides an overview of LLM techniques applied in the domain of stance detection of misinformation.
Section \ref{Section:lm} analyzes supervised, unsupervised, few-shot, and zero-shot learning approaches for stance detection. Section \ref{Section:ma} examines unimodal (signle modality), multimodal (text with images or audio or video), multilingual, and hybrid methods for stance analysis, while Section \ref{Section:ta} explores how stance detection varies based on target types. Section \ref{Section:dataset} summarizes key datasets, evaluation metrics, and benchmarks used in stance detection research, and Section \ref{Section:application} discusses practical uses of stance detection in politics, marketing, healthcare, and social media monitoring. Section \ref{Section:challenges} addresses key challenges and limitations. Finally, Section \ref{future} concludes the survey by synthesizing major insights and outlining promising directions for future research.
This systematic review aims to serve both researchers seeking to advance the field and practitioners looking to implement state-of-the-art stance detection systems. By synthesizing recent advances and identifying key challenges, we provide a roadmap for future research at the intersection of LLMs and stance analysis.

\section{Large Language Models: A Brief Overview}
\label{Section:llm}
\begin{figure*}[!ht]
    \centering
    \captionsetup{justification=justified}   \includegraphics[width=16cm, height=6cm]{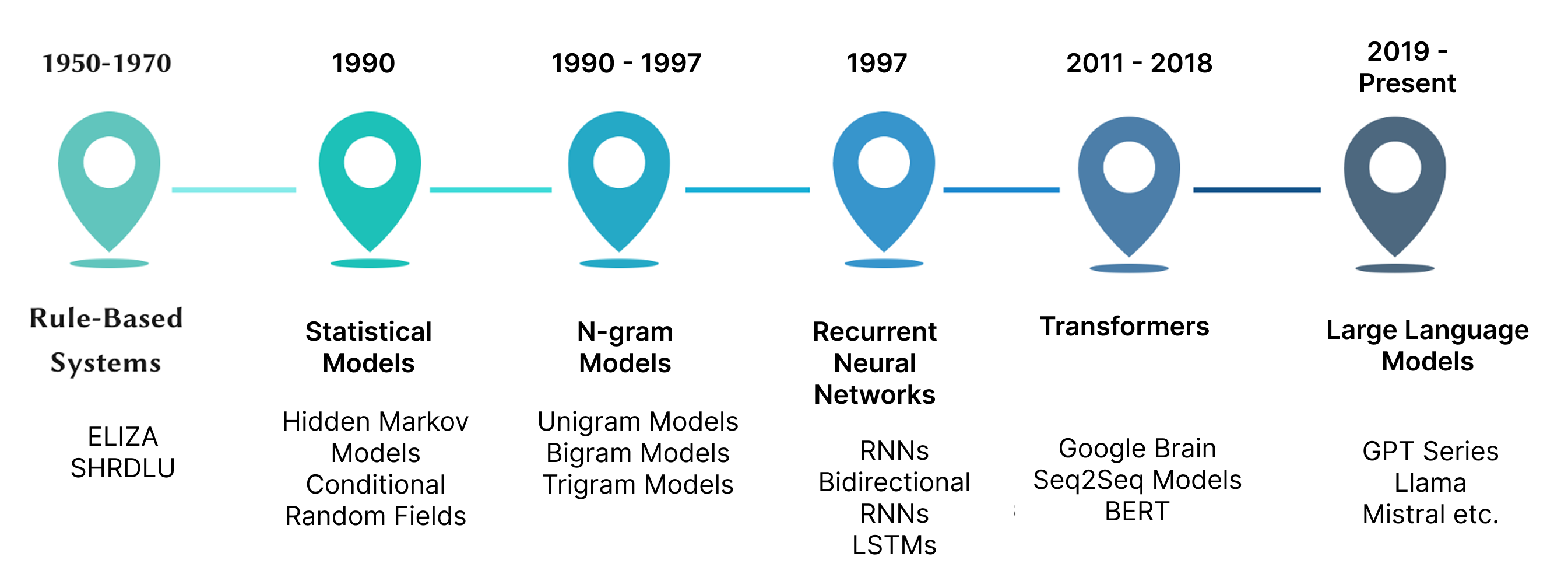}
    \caption{ Evolution of Language Models: From Rule-Based Systems to Modern Large Language Models (LLMs)}
    \label{fig:fig6}
\end{figure*}

Large Language Models (LLMs) are massive neural architectures, specifically generative transformers, trained on large-scale, unlabelled text corpora using self-supervised learning. By learning to predict the next word in a sentence, they internalize complex linguistic structures, semantics, and broad world knowledge. Leveraging transfer learning and prompt-based approaches, LLMs can be adapted to specialized domains and tasks such as summarization, question answering, and dialogue, often with minimal or no task-specific fine-tuning . Figure \ref{fig:fig6} illustrates the taxonomy of the evolution of language models, tracing their development from rule-based systems to modern large language models (LLMs).

\subsection{Evolution and Architectures}
Early language models from 1950–1970 were primarily rule-based systems built on handcrafted grammar. Notable examples include ELIZA (1966) \cite{Weizenbaum_2021}, a pattern-matching chatbot simulating therapeutic dialogue, and SHRDLU (1970) \cite{winograd1971procedures}, which interacted in a limited virtual world, were early landmarks. 
In 1954 IBM-Georgetown machine translation demo \cite{hutchins2005first} helped lay the foundation for NLP, though such systems lacked robustness in ambiguity and context, leading to a shift toward statistical approaches.

The 1990s witnessed the statistical models \cite{church1993introduction} become dominant, using large corpora to identify language patterns. Hidden Markov Models (HMMs) \cite{RABINER_1990} and Conditional Random Fields (CRFs) \cite{lafferty2001conditional} enabled progress in part-of-speech tagging and NER. N-gram models \cite{jelinek1998statistical} improved contextual predictions, paving the way for more sophisticated systems.

The emergence of neural models gained traction with LSTMs \cite{6795963}, capturing long-range dependencies, followed by GRUs \cite{Cho_2014}, offering simplified performance. The development of integrated frameworks such as CoreNLP \cite{Manning_2014} in 2010  further expanded access to NLP tasks like sentiment analysis. These advances set the stage for today’s LLMs.
Google Brain's word2vec \cite{mikolov2013efficient} in 2011 introduced word embeddings, improving contextual understanding. Seq2Seq models with attention \cite{sutskever2014sequence} advanced tasks like translation. A major leap came with the Transformer architecture \cite{vaswani2017attention} introduced in 2017, which replaced recurrence with self-attention, revolutionizing model scalability and contextual learning.
BERT \cite{devlin2019bert} applied bidirectional transformers and pretraining to capture nuanced word relationships, setting new NLP benchmarks and enabling fine-tuning across diverse tasks in 2019. OpenAI’s GPT series further expanded LLM capabilities: GPT-2 \cite{radford2019language} in 2019 showed the strength of unsupervised generation; In 2020, GPT3  \cite{brown2020language} popularized few- and zero-shot learning \cite{10888887}; and GPT4 \cite{achiam2023gpt} introduced multimodal reasoning and controllability in 2020.
In the open-source space, Meta’s LLaMA \cite{touvron2023llama} in 2023 democratized LLM access, emphasizing customizability and responsible AI. Zephyr-7B \cite{tunstall2023zephyr}, a fine-tuned Mistral-7B variant, uses distilled direct preference optimization \cite{yue2024distilling} and AI Feedback to surpass LLaMA2-Chat-70B, aligning closely with user intent, all with minimal training and no human annotation.

\subsection{Pretraining and Fine-tuning Paradigms}
Large language models (LLMs) are pretrained using Causal Language Modelling (CLM) and Masked Language Modelling (MLM). CLM, used in models like GPT-2 \cite{radford2019language} and GPT-3 \cite{brown2020language}, predicts each token based on previous ones, making it ideal for generation tasks but limited by unidirectional context. In contrast, MLM, as used in BERT \cite{devlin2019bert}, masks input tokens and predicts them using both left and right context, enabling deep contextual understanding for tasks like classification and QA. Enhanced variants such as RoBERTa \cite{liu2019roberta}, SpanBERT \cite{Joshi_2020}, and ELECTRA \cite{clark2020electra} improve on MLM with dynamic masking and more efficient prediction.
Pretraining effectiveness hinges on high-quality, diverse, and filtered data \cite{gao2020pile, raffel2020exploring} to ensure generalization and safe deployment. Finetuning further adapts pretrained LLMs to specific tasks or domains \cite{Gururangan_2020}, combining general knowledge with task-specific insights.
To improve alignment with human intent, Reinforcement Learning with Human Feedback (RLHF) \cite{NIPS2017_d5e2c0ad} uses human-ranked outputs to train a reward model, guiding generation. Direct Preference Optimization (DPO) \cite{NEURIPS2023_a85b405e} simplifies this by directly optimizing preference via contrastive loss, while AI Feedback (AIF) automates ranking using LLMs, removing the need for manual labelling.
Parameter-Efficient Fine-Tuning (PEFT) methods like LoRA \cite{hu2022lora}, Adapters \cite{houlsby2019parameter}, and Prefix-Tuning \cite{Li_prefix_2021} reduce resource demands by modifying only a small portion of the model, enabling efficient and flexible adaptation for various applications.

\subsection{In-context Learning, Prompt Engineering, and Instruction Tuning}

In-Context Learning (ICL) enables models to perform new tasks by conditioning on examples within the prompt, without updating model weights. By observing demonstrations, such as Q\&A pairs or completions, models like GPT-3 generalize to similar tasks in zero-shot, one-shot, or few-shot settings. This makes ICL especially useful for rapid task adaptation, low-resource scenarios, and domains with limited labelled data.
Prompt Engineering is the practice of crafting inputs to guide model behaviour without fine-tuning. Small changes in phrasing, structure, or formatting can significantly affect output quality \cite{li-etal-2025-mitigating-biases, ZHANG2025103214, 10.1145/3701716.3715465}. Effective prompts enhance performance in tasks like summarization, reasoning, and translation, particularly in zero and few-shot contexts \cite{LUO2026131178}.
Instruction Tuning trains models on datasets where tasks are described in natural language with example completions, teaching them to follow explicit instructions. This improves generalization, reduces hallucinations, and boosts performance across a wide range of tasks. Models like FLAN \cite{JMLR:v25:23-0870} and InstructGPT \cite{NEURIPS2022_b1efde53} exemplify this approach, showing strong zero- and few-shot capabilities aligned with user intent.

\subsection{Performance in Downstream Tasks}
\begin{figure*}[!ht]
    \centering
    \captionsetup{justification=justified}   \includegraphics[width=15cm, height=8.5cm]{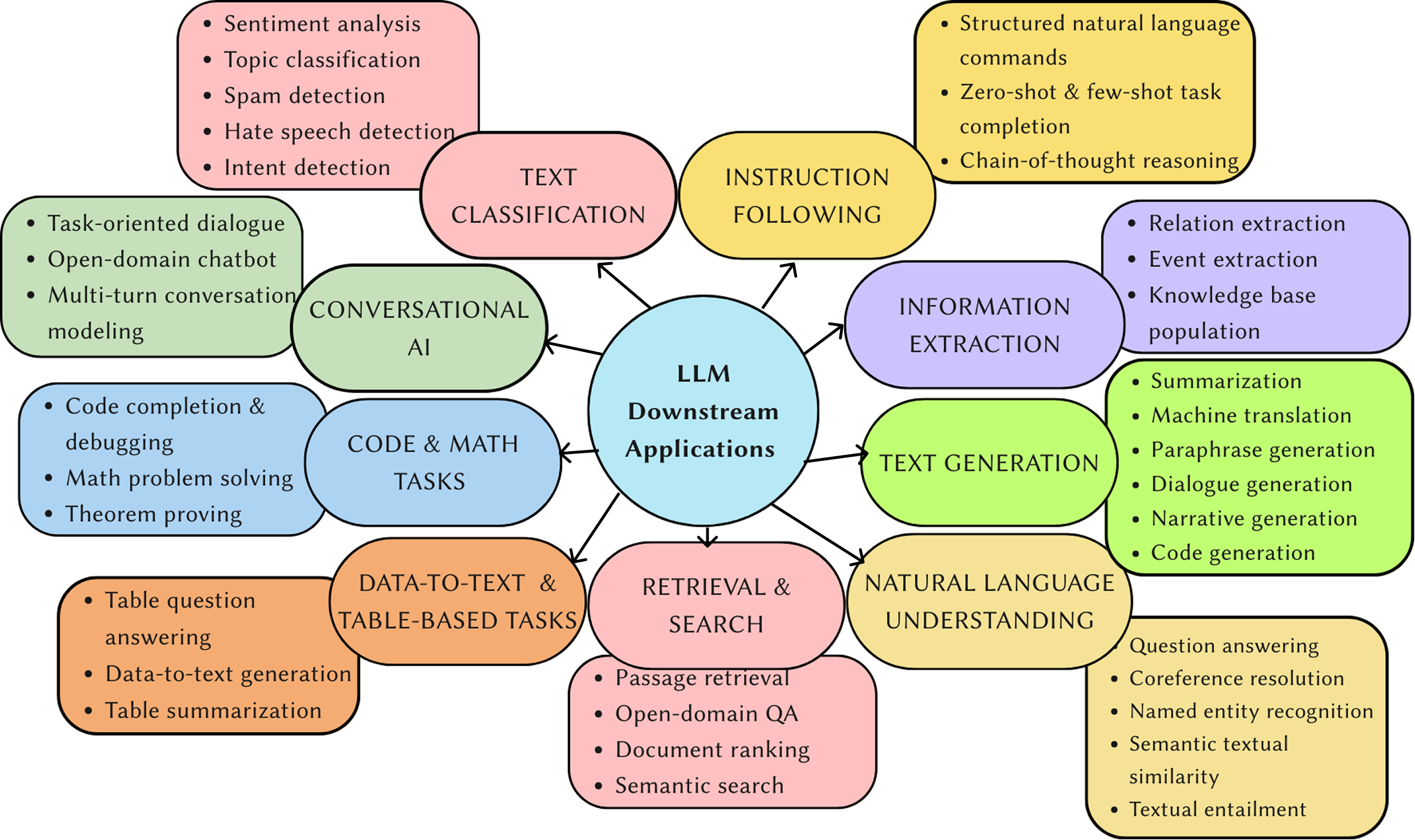}
    \caption{Downstream Applications of LLMs}
    \label{fig:fig7}
\end{figure*}

LLMs have been used in various downstream tasks, some of which have been listed in Figure~\ref{fig:fig7}. 
Text classification is a key downstream task for LLMs, involving the assignment of category labels to input text. Applications include sentiment analysis, hate speech detection, spam filtering, topic classification, and intent recognition. LLMs leverage pretrained contextual knowledge to outperform traditional models, especially in low-resource or noisy settings. Their generalization ability enables accurate performance across domains and languages, even in zero- or few-shot scenarios. 
LLMs excel at extracting structured information from unstructured text. Tasks like Named Entity Recognition (NER), relation and event extraction, textual entailment, and coreference resolution benefit from their deep semantic understanding. Pretrained models like BERT achieve strong performance, especially when fine-tuned on annotated data. Instruction tuning and domain-adaptive pretraining further enhance their accuracy across diverse languages and specialized domains.
LLMs power key generative tasks like summarization, translation, and dialogue generation. They produce abstractive or extractive summaries, support multilingual translation (e.g., mT5), and serve as core engines in chatbots. Their ability to generate long-form content also enables creative applications like storytelling and instructional writing.
Question answering (QA) tasks leverage LLMs to extract or generate answers from text, including complex forms like multi-hop QA. Retrieval-Augmented Generation (RAG) enhances accuracy by combining LLMs with retrieval systems, making them effective for open-domain and knowledge-intensive QA.
LLMs exhibit emerging strengths in reasoning tasks, including commonsense, symbolic, and arithmetic reasoning. Techniques like chain-of-thought prompting enhance multi-step reasoning, while few-shot and in-context learning enable adaptation to planning and goal-oriented tasks with minimal examples.

\section{Learning Methods}\label{Section:lm}
\begin{figure*}[!ht]
    \centering
    \captionsetup{justification=justified}   \includegraphics[scale=0.5]{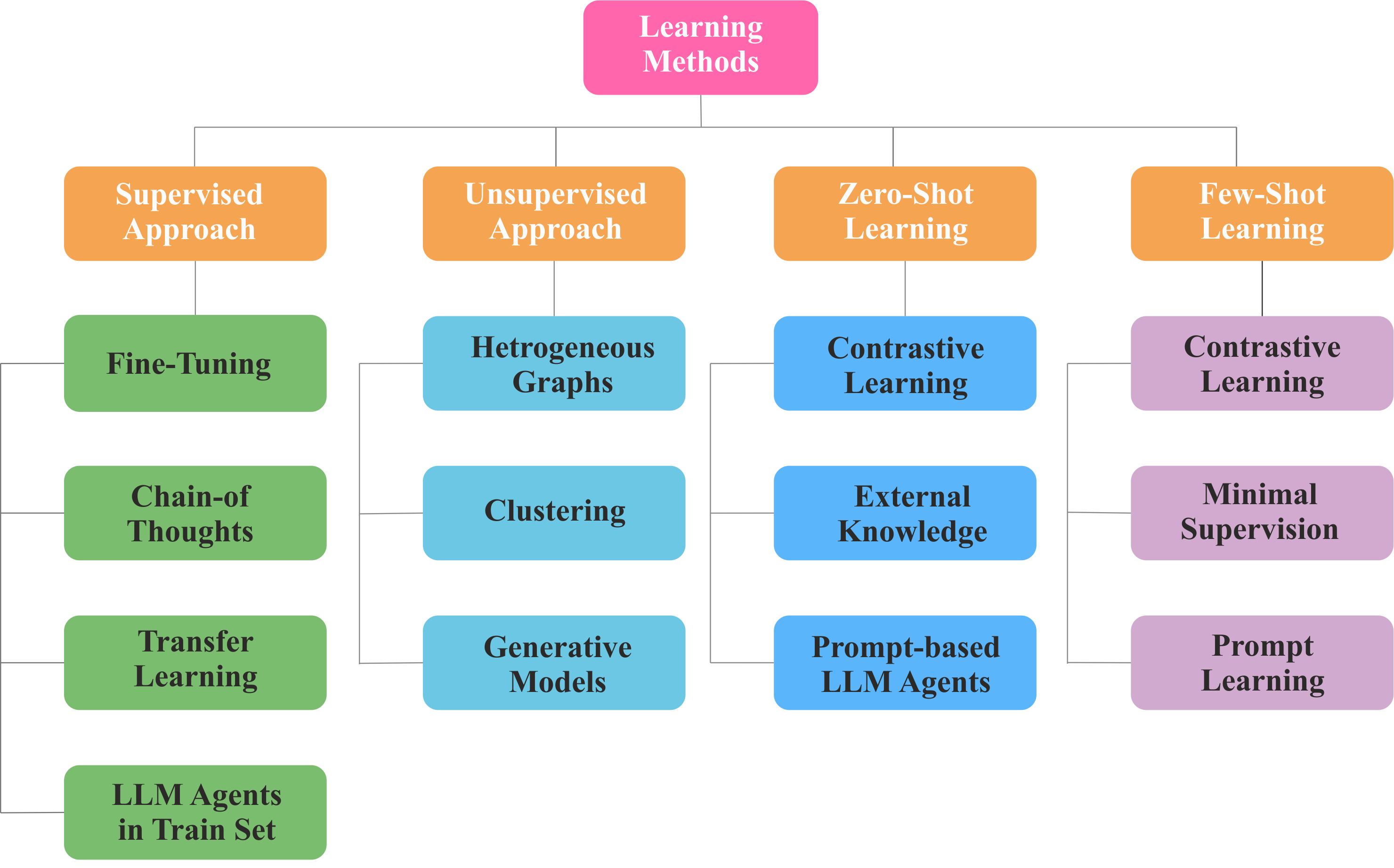}
    \caption{Learning Methods in Stance Detection}
    \label{fig:fig8}
\end{figure*}

This section explores the various learning methods used in stance detection, highlighting key approaches such as supervised, unsupervised, zero-shot, and few-shot learning. Each of these techniques leverages advancements in natural language processing and machine learning to detect stances in social media and user-generated content. From traditional supervised models that rely on labelled data to more flexible, unsupervised, and few and zero-shot methods that require minimal or no supervision, these approaches offer scalable and adaptable solutions. 
As shown in Figure~\ref{fig:fig8}, these methods are classified based on their underlying approaches and techniques.
The following content provides a comprehensive overview of these methods and their contributions to improving stance detection across diverse domains.

\subsection{Supervised Approach}

\begin{figure*}[!ht]
    \centering
    \captionsetup{justification=justified}
    \includegraphics[width=\textwidth]{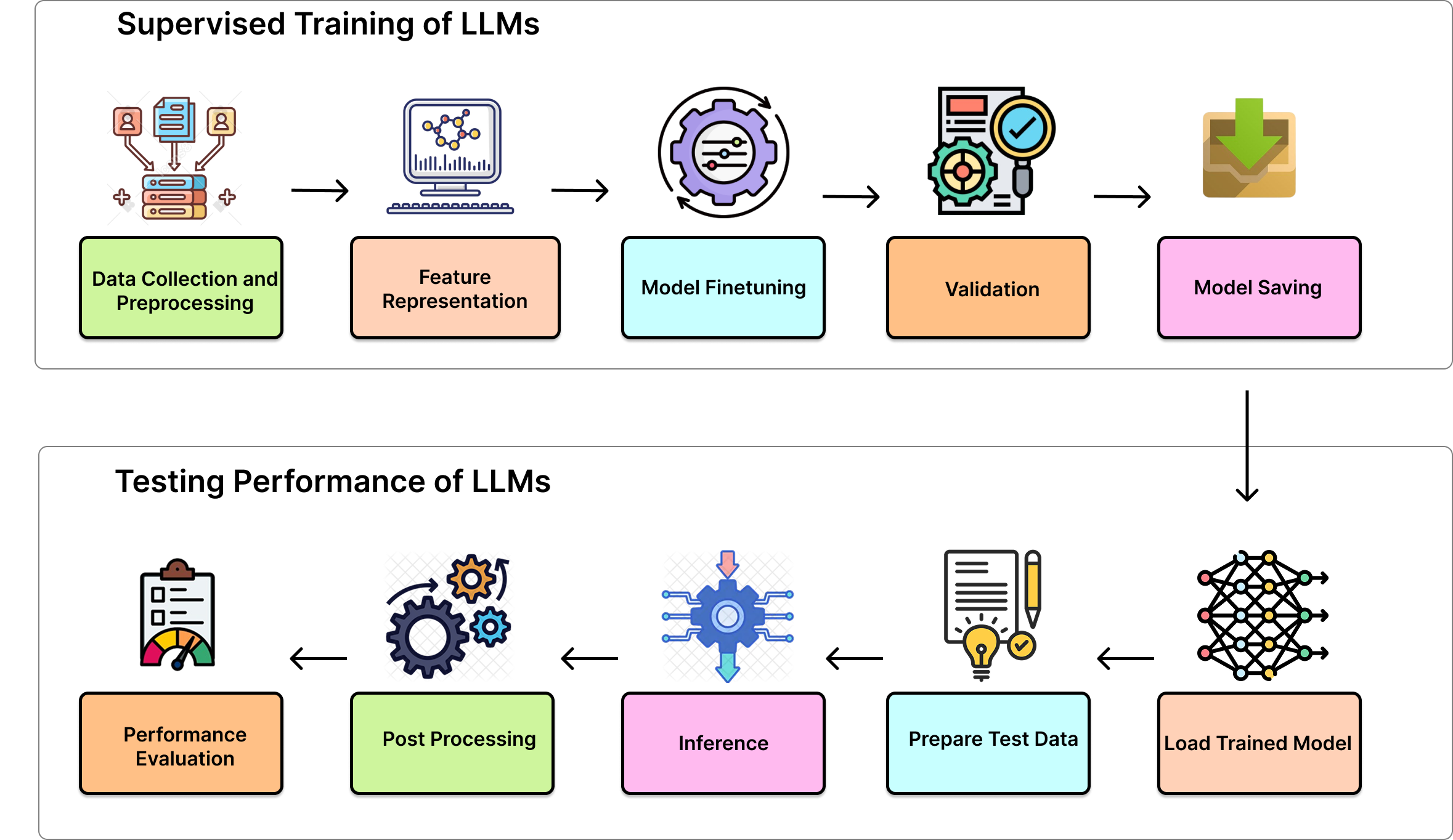}
    \caption{Supervised Training and Testing of LLMs}
    \label{fig:fig9}
\end{figure*}

Supervised learning is the most widely used approach in stance detection, where models are trained on labelled datasets containing texts annotated with stance labels. 
This approach leverages large-scale labelled data to learn discriminative features that map input texts to their respective stance classes. The supervised learning process is illustrated in Figure ~\ref{fig:fig9}. Recent advancements in deep learning, particularly pre-trained language models (PLMs), have significantly improved the performance of supervised stance detection.
Fu \textit{et al.} \cite{FU2022108657} proposes a supervised stance detection framework that incorporates opinion-towards information to enhance classification accuracy. The authors utilize BERT (Bidirectional Encoder Representations from Transformers) to generate contextualized embeddings, which effectively capture stance-related signals in social media texts. By fine-tuning BERT on stance-labelled data, their model demonstrates improved generalization across diverse topics and linguistic styles.
Gatto \textit{et al.} \cite{gatto-etal-2023-chain} introduces a supervised approach for stance detection, leveraging chain-of-thought embeddings to capture the reasoning behind stance shifts in social media texts.
For multimodal stance detection, Niu \textit{et al.} \cite{Niu_2024} introduces a supervised approach for stance detection in multi-turn conversations, utilizing both a textual encoder and a visual encoder to handle textual and visual data. The model uses a LLaMA-2 model \cite{touvron2023llama2openfoundation}, LoRA (Low-Rank Adaptation) \cite{von-platen-etal-2022-diffusers}, and prompts for efficient fine-tuning, allowing it to process image captions and other multimodal content effectively.
In scenarios involving multiple targets, Liu \textit{et al.} \cite{LIU2024108515} proposes a supervised comparative learning framework that detects stance agreement across multiple targets by comparing stance representations between different target-oriented texts. The approach models agreement or disagreement patterns between them using a contrastive learning approach.
Wang \textit{et al.} \cite{wang-etal-2024-deem} presents a supervised stance detection framework where the DEEM model leverages multiple expert models, each curated from subsets of the training data based on stance-relevant characteristics such as topic, sentiment, or linguistic patterns. These specialized experts are trained to capture diverse stance expressions, and a dynamic gating mechanism selects and combines their outputs based on the input instance.
Collectively, these supervised approaches demonstrate the versatility of modern large language model techniques in stance detection, from leveraging pretrained models such as BERT to  ChatGPT for modelling complex stance dynamics in conversations.

Collectively, in the field of stance classification, these studies underscore the versatility and efficacy of supervised learning in the field, with PLMs serving as the foundation for capturing nuanced stance-related features. While each method excels in specific contexts, such as target-aware modelling for structured debates or interpretable reasoning for social media, the integration of multimodal data and dynamic expert ensembles highlights the field's trajectory toward increasingly robust and adaptive systems. Future research should focus on unifying these approaches to address remaining challenges, such as generalizability across domains and real-time adaptation to evolving discourse. By refining these methodologies, the next generation of stance detection systems will further bridge the gap between theoretical innovation and practical application, enabling more accurate and interpretable stance analysis in diverse communicative contexts.

\subsection{Unsupervised Approach}
Unsupervised learning for stance detection aims to identify stance expressed in a text without relying on annotated training data. Instead of learning from labelled examples, it leverages various techniques such as clustering approaches \cite{SINGH2025127292}, inheriting patterns in language, such as semantic roles, sentiment, and discourse structures, to infer stance.
Kobbe \textit{et al.} \cite{kobbe-etal-2020-unsupervised} leverages linguistic and semantic cues to identify stance without requiring labelled data. It detects stance by analyzing the causal relation between actions and their perceived good or bad consequences using handcrafted rules and external lexical resources.
Pick \textit{et al.} \cite{Pick_Kozhukhov_Vilenchik_Tsur_2022} constructs a heterogeneous graph where nodes represent posts, users, and threads, and edges capture their interactions, such as replies or authorship. This graph is then used to learn unsupervised structural embeddings that capture stance-indicative patterns based on the network structure of conversations and user interactions.
Abeysinghe \textit{et al.} \cite{9835637} employ Sentence Transformers to generate semantic embeddings of tweets, capturing their contextual meanings. By comparing these embeddings to those of the target web articles, the method assesses the similarity or divergence in content. Subsequently, clustering techniques are applied to group users based on the semantic alignment of their tweets with the articles.
Kuo \textit{et al.} \cite{10.1145/3589335.3651467} uses an unsupervised clustering approach to detect political stances of fan pages without relying on labelled data. It groups fan pages based on similarities in their content and behaviour, allowing the identification of stance patterns through the emergent structure of the data rather than predefined categories.
Samih \textit{et al.} \cite{samih-darwish-2021-topical} introduces an unsupervised approach to determine user stances on controversial topics by leveraging their Twitter activity. It performs unsupervised classification by clustering users based on similarities in their tweet content.
Zhao \textit{et al.} \cite{ZHAO2024102386} presents an unsupervised approach to stance detection that does not rely on labelled data for new or unseen topics. Instead, it focuses on learning transferable features from existing data to generalize to novel targets.
Overall, unsupervised stance detection uses the underlying structures within language, content, and user interactions to uncover stance-related patterns. By eliminating the dependency on labelled data, these approaches provide scalable, adaptable, and domain-independent solutions for real-world stance detection tasks.

From the linguistic pattern analysis to the semantic embedding clustering and the transferable feature learning, the unsupervised methods presented here show how structural, semantic, and network-based patterns can efficiently reveal stance without the need for human annotations.  Supervised and unsupervised methods form complementary foundations of contemporary stance detection research. While unsupervised approaches offer scalability and domain flexibility, supervised approaches provide precision through learned discriminative features.  Subsequent growth in the field of stance detection ought to concentrate on hybrid frameworks that integrate the advantages of both strategies, possibly via adaptive knowledge transfer or semi-supervised learning.

\subsection{Zero-Shot Learning}
Recent advances in large language models (LLMs) and transformer-based architectures have enabled effective zero-shot stance detection without the need for task-specific training. By leveraging pre-trained models such as BART, RoBERTa, and GPT variants, researchers frame stance detection as a natural language inference (NLI) or masked language modelling (MLM) task. These methods harness the generalization capabilities of LLMs to handle diverse topics, languages, and data conditions with zero supervision, making them especially suited for emerging or under-resourced domains.

Advances in zero-shot stance detection (ZSSD) focus on combining semantic structuring, contrastive learning, and external knowledge to enhance generalization to unseen targets. Zhao \textit{et al.} \cite{Zhao_2024} introduce a hierarchical contrastive framework using an Aspect Feature Mapper (AFM) to project inputs into semantic subspaces, with inter- and intra-aspect contrastive learning capturing diverse distinctions. A retrieval-based attention mechanism aligns multi-aspect features with target-utterance representations for improved stance detection. Liang \textit{et al.} \cite{Liang_2022} employ supervised contrastive learning with BERT encodings and data augmentation to extract target-invariant features, which are fused with target-specific features via attention for stance prediction on unseen targets.
Zhu \textit{et al.}\cite{Zhu_2022} improve generalization by injecting Wikipedia-based knowledge into BERT to bridge semantic gaps. Similarly, Zou \textit{et al.} \cite{Zou_2022} use a self-supervised task with LDA-based masking and hierarchical contrastive learning to disentangle and refine transferable stance features. Wang \textit{et al.} \cite{Wang_2024} propose a joint model- and data-centric approach using BART for target-aware keyphrase generation and topic-level attention. Stance prediction is reframed as text entailment with meta-learning for target adaptation and generalization, supported by supervised contrastive loss for robust representations.
Lin et al. \cite{lin-etal-2024-kpatch} propose a two-stage knowledge injection method combining knowledge compression and task guidance. Subgraphs from external knowledge graphs are encoded into compact matrices, which are then used to guide PLMs in retrieving relevant knowledge for tweet-topic pairs. The PLM is fine-tuned for final stance prediction.
Guo et al. \cite{Guo_2024} use role-playing LLM agents to generate topic explanations and pseudo-samples for unseen targets. These are integrated into BERT via a refined input format capturing the relationship between text, topic, and explanation, followed by a fully connected layer for stance prediction. Zhao et al. \cite{zhao-etal-2024-zerostance} propose CHATStance, a synthetic open-domain dataset generated with ChatGPT. Claims are derived from Kialo categories and geographically contextualized prompts, with stance-labelled texts created by ChatGPT. Variability-based filtering removes trivial and noisy instances, enhancing generalization for open-domain ZSSD.
Collectively, these studies underscore a converging trend in ZSSD research: the strategic combination of representation disentanglement, supervised or self-supervised contrastive objectives, meta-learning strategies, and dynamic knowledge augmentation—whether from external graphs, Wikipedia, or large language models—to construct stance detectors that are adaptable, semantically aware, and robust across diverse, unseen targets.

The surveyed methods, from contrastive learning frameworks to knowledge-enhanced architectures, and LLM-generated synthetic data demonstrate how strategic combinations of representation learning, external knowledge integration, and meta-learning can overcome the limitations of unseen targets. These approaches collectively establish three critical advancements: (1) semantic disentanglement through hierarchical contrastive objectives, (2) dynamic knowledge adaptation via graphs or LLMs, and (3) robust generalization through synthetic data augmentation. While current methods excel in open-domain adaptability, upcoming advancements must address challenges in cross-lingual transfer, multimodal stance interpretation, and computational efficiency to fully realize zero-shot detection's potential for real-time, large-scale discourse analysis.

\subsection{Few-Shot Learning}
This section focuses on stance detection strategies in low-resource settings. By generalizing to new targets or domains using limited or no supervision, these techniques enable scalable and adaptable stance classification across diverse topics and are particularly valuable for real-time applications where rapid adaptation to new topics is essential.
Few-shot stance detection has emerged as a promising solution to the reliance on large annotated datasets, which are often scarce or costly to obtain for emerging or domain-specific topics. 
Recent advancements in stance detection reflect a clear shift from rigid, fully-supervised pipelines to flexible, inference-driven architectures that leverage prompt learning, external knowledge, and pre-trained language models. The FSRD framework by Li \textit{et al.}\cite{Li_2024} exemplifies this evolution by introducing a few-shot fuzzy rumor detection system using stance-enhanced prompt learning. Tailored for low-resource settings, it transforms rumor texts into interrogative template pairs for BERT-based NSP tasks, enabling effective rumor classification with minimal supervision. FSRD aligns closely with recent few-shot stance detection approaches.
A notable advancement in few-shot stance detection is the commonsense-enhanced model by Liu \textit{et al.} \cite{Liu_2021}, which incorporates both structural and semantic relational knowledge from ConceptNet. The model encodes document-topic pairs using BERT and maps phrases to associated concepts in ConceptNet. 
Building on structural knowledge paradigms, the approach by Jiang \textit{et al.} \cite{Jiang_2022} reframes stance detection as a masked prompt-based learning task using target-aware verbalizers and sequential distillation. ExPrompt \cite{Qudus_2024} advances few-shot prompting through in-context learning, using example-based templates such as Supports, Refutes to prompt LLMs such as Llama-3 and Mixtral without fine-tuning, establishing strong few-shot baselines. Similarly, Giddaluru \textit{et al.} \cite{Giddaluru_2024} show that models such as BART-large-MNLI can generalize to code-mixed, informal text with minimal supervision, underscoring the adaptability of prompt-based inference.
At the semantic frontier, SentKB-BERT \cite{Hardalov_2022} enhances stance representation through sentiment-aware filtering and integration of external knowledge from ConceptNet and Wikipedia. It continues the tradition of earlier frameworks that employed relational subgraph reasoning, prompt engineering, and latent topic modelling to uncover implicit links and bridge semantic gaps under few-shot regimes. This builds on prior work, a prompt-based and contrastive learning framework by Liu \textit{et al.} \cite{Liu_2022}. The latter also introduced latent topic modelling to align stance-consistent examples, uncovering implicit semantic links between targets in few-shot settings.

In conclusion, the shift from knowledge-grounded models such as CompGCN to prompt-based and contrastive approaches marks a move toward flexible, inference-driven stance detection. By combining external knowledge, prompt engineering, and PLM-based few-shot reasoning, modern systems achieve robust generalization across domains, languages, and data regimes, moving beyond rigid, fully-supervised paradigms.

\section{Approaches}\label{Section:ma}

The field of stance detection has undergone a paradigm shift with the emergence of large language models, fundamentally transforming how computational systems analyze and interpret stance.  
A key strength of LLMs in stance detection lies in their adaptability and transfer learning capabilities. Through fine-tuning on domain-specific datasets, these models can be optimized for particular applications while retaining their broad linguistic knowledge. 

\begin{figure*}[!ht]
    \centering
    \captionsetup{justification=justified}   \includegraphics[width=15cm, height=8.5cm]{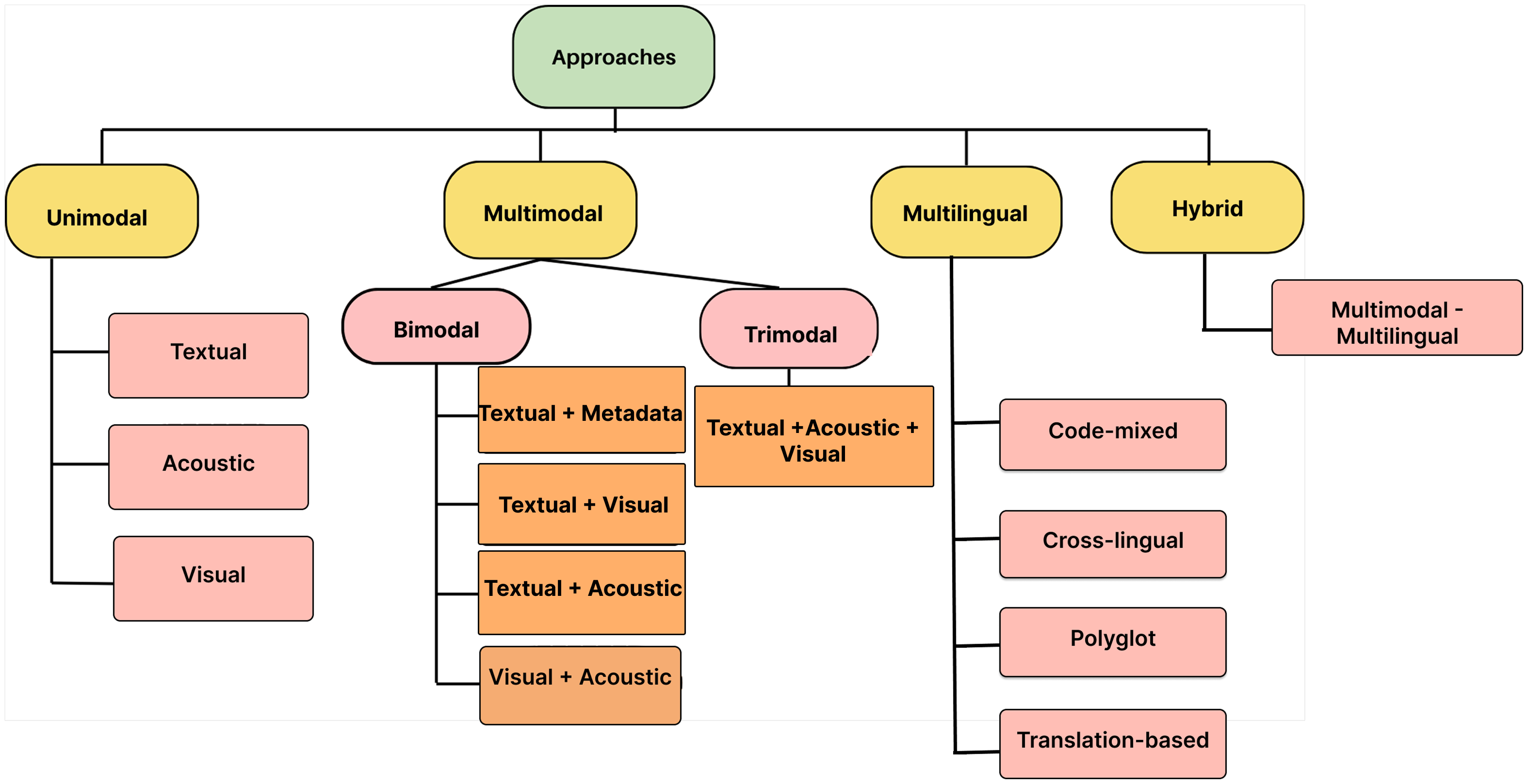}
    \caption{LLM Approaches}
    \label{fig:fig10}
\end{figure*}

The versatility of LLMs has given rise to three primary methodological approaches in stance detection research. First, we explore unimodal approaches, which rely on a single data modality to infer stance. Next, we discuss multimodal approaches, where multiple data modalities such as text, images, or metadata are integrated to enhance the understanding of stance. Finally, we consider hybrid approaches, which combine traditional machine learning techniques with deep learning or fuse domain knowledge with data-driven methods to leverage the strengths of both paradigms, all listed in Figure~\ref{fig:fig10}. 

\subsection{Unimodal Approaches}
Unimodal approaches focus on leveraging a single modality of data, especially text and images, and have various applications \cite{RAGHAW2024108821, RAGHAW2025107363,DAR2024112526}. With the evolution of NLP, unimodal methods have advanced from early machine learning classifiers to deep learning techniques and now to large pre-trained transformer-based language models.
Building upon the foundations, recent studies have adopted advanced techniques that employ knowledge enhancement, prompt-based fine-tuning, and large pre-trained models such as BERT and GPT.
For instance,
Kawintiranon \textit{et al.} \cite{kawintiranon-singh-2021-knowledge} utilizes a BERT-based masked language model fine-tuned with a weighted log-odds-ratio to identify words that are highly indicative of stance. An attention mechanism is then applied to focus on these key words, enhancing the model's ability to detect stances in short and informal tweets.
Sun \textit{et al.} \cite{10.1007/978-3-031-30678-5_18} enriches the textual input with external knowledge from knowledge graphs to improve understanding of the relationship between a target and the surrounding text.
Jiang \textit{et al.} \cite{10.1145/3477495.3531979} adopts prompt-based fine-tuning of BERT for stance classification, using target-aware prompts that integrate the target into the input text to enhance the model’s comprehension of stance-target relations.
Li \textit{et al.} \cite{li-etal-2023-new} aims to extract both the target and the corresponding stance from social media texts where the target is often implicit or unknown. The proposed framework first identifies potential targets using either classification or generation methods and then detects the stance towards the predicted target. 
Ma \textit{et al.} \cite{MA2024103528} emphasizes textual information as the primary source for both stance detection and veracity prediction. It utilizes the textual content of user posts in the conversation thread, capturing semantic relationships and stance patterns, thereby improving the model’s understanding of how different stances contribute to rumor verification.
Building on these developments, recent advancements have increasingly embraced prompt-based methods using large pre-trained language models such as GPT-3. These models not only support flexible, instruction-driven input processing but also allow the integration of external knowledge and situational cues directly through natural language prompts. This makes them especially effective in open-domain and implicit target scenarios.
Li \textit{et al.} \cite{li-etal-2023-stance} leverages ChatGPT to augment stance detection by integrating relevant external knowledge into the input text. It reformulates social media posts by incorporating contextual background information retrieved via ChatGPT, thereby enriching the textual input and enabling the model to make more informed and accurate stance predictions.
Zhao \textit{et al.} \cite{zhao-etal-2024-zerostance} employs ChatGPT to generate synthetic stance-labelled data across various domains, facilitating open-domain stance detection in zero-shot scenarios without the need for manual annotation. Similarly, Zhang \textit{et al.} \cite{zhang-etal-2024-llm-driven} introduce a framework that enriches textual input with external knowledge using LLMs, improving performance on zero-shot and cross-target tasks by enhancing contextual comprehension. Complementing these approaches, Lan \textit{et al.} \cite{Lan_Gao_Jin_Li_2024} proposes a multi-agent architecture where each LLM-based agent assumes a specific role to collaboratively process textual data. Collectively, these works highlight the growing effectiveness of LLM-driven unimodal models in capturing subtle and nuanced stance information from text.
These advancements underscore a clear shift from rigid, feature-based models toward more adaptable, knowledge-enriched, and generalizable unimodal stance detection systems powered by large language models.
Overall, unimodal approaches demonstrated impressive progress in understanding stance. However, challenges remain, including difficulties in handling implicit targets, achieving robust domain generalization, and compensating for the lack of multimodal cues. These limitations pave the way for the development of more comprehensive multimodal or hybrid approaches that can further advance stance detection capabilities.

\subsection{Multimodal Approaches}

\begin{figure*}[!t]
    \centering
    \captionsetup{justification=justified}
    \includegraphics[width=\textwidth]{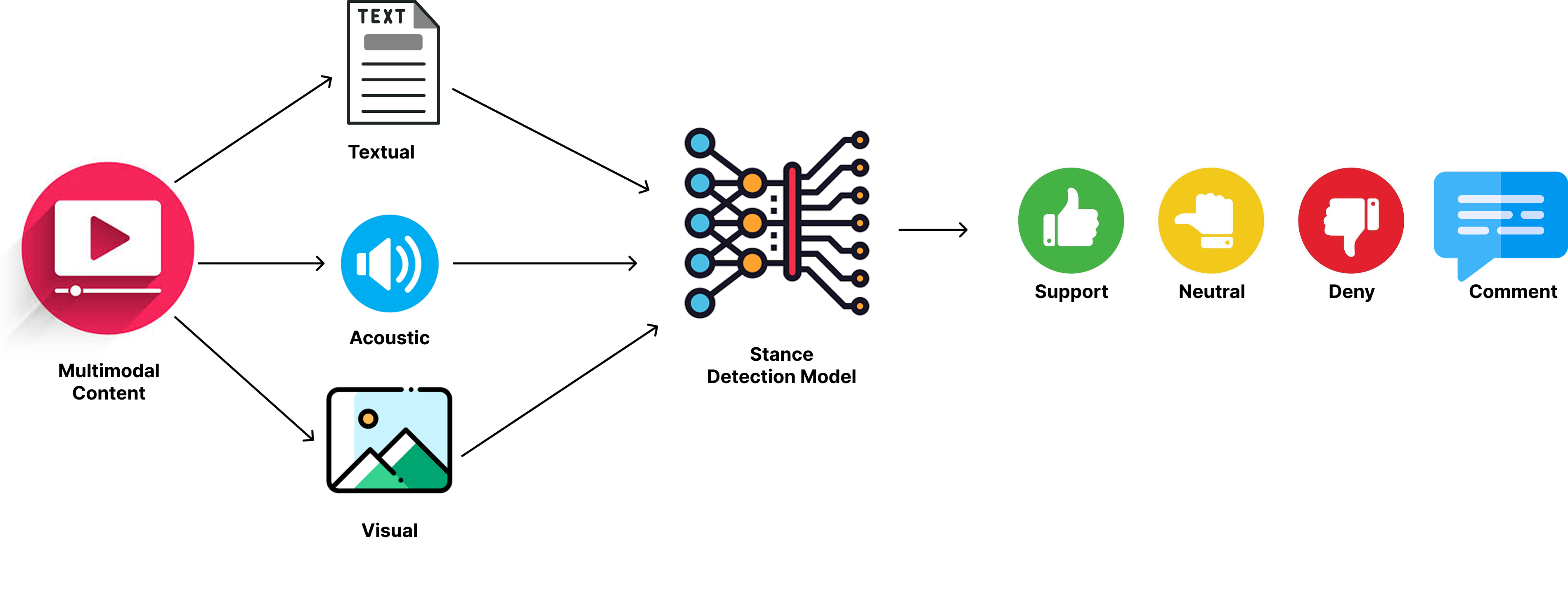}
    \caption{Multimodal Stance Detection}
    \label{fig:fig11}
\end{figure*}

With the growing popularity of social media and video-sharing platforms, online content has become increasingly diverse and complex, often combining multiple modalities such as text, audio, and images in formats such as memes, reels, and videos \cite{DAR2025125337, BANSAL2024109417, 9806458}. These multimodal forms of communication frequently convey implicit opinions not only through language but also through visual and auditory cues. This introduces significant challenges for stance detection, as it requires aligning and interpreting semantically heterogeneous signals across modalities. Traditional models often struggle to effectively capture both intra-modal and inter-modal interactions. In contrast, recent advances in transformer-based architectures and large language models (LLMs) have shown strong potential in modelling cross-modal dependencies. Their ability to leverage self-attention mechanisms and transfer learning makes them well-suited for stance inference from complex multimodal content. As shown in Figure~\ref{fig:fig11}, the figure illustrates the multimodal approach to stance detection, which combines multiple modalities such as text, audio, and visual features to enhance stance classification accuracy.
Recent advancements in multimodal stance detection (MSD) reflect a shift from simple text-image pair analysis toward more context-aware and semantically aligned frameworks. For instance, Chen \textit{et al.} \cite{Niu_2024} introduce the MmMtCSD dataset and propose MLLM-SD, a transformer-based architecture designed to jointly learn stance representations from text and images within multi-turn conversations, acknowledging the complex dynamics of real-world online discourse. Similarly, in the context of the 2021 Taiwanese referendum, Kuo \textit{et al.} \cite{Kuo_2024} integrate textual, visual, and social features—including metadata, sentiment, and user interactions—using S-BERT for text, BEiT for images, and GraphSAGE for inductive graph learning which substantially improves stance detection and visualization by capturing structural and contextual relationships among politically affiliated posts.

Complementing these, SERN \cite{Xie_2021} enhances fake news detection by implicitly extracting stance from post-reply pairs via multimodal fusion. It combines BERT and ResNet-152 features using sentence-guided visual attention and a fully connected graph \cite{mahmud-etal-2025-grasp} for stance propagation, demonstrating how attention-based multimodal modelling strengthens stance extraction and fact-checking.
Thapa \textit{et al.} \cite{thapa-etal-2025-multimodal} study multimodal stance detection in socio-political memes by jointly modelling text and images.
Their shared task shows that multimodal models clearly outperform text-only baselines.
However, stance classification remains challenging due to sarcasm, cultural context, and label imbalance.
Barel \textit{et al.} \cite{barel-etal-2025-acquired} propose a multimodal stance detection framework that combines textual content with conversational structure.
Their model fuses transformer-based text embeddings with social interaction graphs to capture stance more accurately.
Results show that adding structural context significantly improves stance prediction over text-only models.
Expanding the scope of MSD to few-shot and cross-target settings, CT-TN \cite{Khiabani_2024} addresses few-shot and cross-target stance detection by combining RoBERTa-based post-topic embeddings with graph-based classifiers over user followers, friends, and likes. Using majority voting, it fuses textual and social features to enhance generalization through both content and relational cues. Adding to this trend, Barel \textit{et al.} \cite{barel2024acquired} propose a multimodal stance detection model combining Sentence-BERT embeddings with speaker interaction graphs via a Gated Residual Network. By fusing semantic and structural cues, it predicts utterance- and speaker-level stances, performing well on small, noisy datasets where speaker dynamics matter.

Extending this multimodal stance paradigm further, Yao 
\textit{et al.}\cite{Yao_2023} present an end-to-end fact-checking and explanation framework that verifies input claims using large-scale web sources (articles, images, videos, and tweets) and generates explanations. It integrates SBERT, CLIP, BERT, and BART for similarity, cross-modal representation, retrieval, and reasoning, highlighting stance detection’s role in interpretable, evidence-based fact-checking.

Recent developments in multimodal stance detection (MSD) show that combining social, visual, and textual cues greatly improves classification performance and robustness.  While attention methods enhance semantic alignment between modalities, transformer-based architectures and graph-augmented models efficiently capture cross-modal interactions.  The three primary tendencies that emerge from these methods are (1) context-aware fusion (such as social networks and multi-turn discussions), (2) generalization across targets (few-shot, cross-domain), and (3) explainability (fact-checking with evidence retrieval).  Scaling to real-time data, cutting computational costs, and enhancing cross-lingual/modal transfer are still difficult tasks, nevertheless.  To close the gap between lab performance and practical implementation, future research should concentrate on self-supervised multimodal pretraining, lightweight fusion approaches, and human-in-the-loop interpretability.

\subsection{Multilingual Approaches}
In the NLP domain, multilingual approaches have emerged as effective solutions for enabling models to perform across diverse languages with limited annotated data \cite{REHMAN2025126285, ZIAURREHMAN2023103450, 10.1145/3618057}.
Recent advancements in stance detection address low-resource and cross-lingual challenges using transformer-based models and knowledge transfer. Jaziriyan \textit{et al.} \cite{Jaziriyan_2021} advances Arabic stance detection by dynamically identifying debated targets in tweets, diverging from fixed-target methods. Using Arabic BERT, they explore three architectures: fine-tuning [CLS] embeddings, combining [CLS] with average-pooled embeddings, and feeding frozen embeddings into a BiLSTM. For Persian,  Nasiri \textit{et al.} \cite{Nasiri_2022} employ ParsBERT and Easy Data Augmentation (EDA) to overcome data scarcity. The model fine-tunes a linear classifier for four-way classification using data from Shayeaat and Fakenews, consisting of claim-headline and claim-body pairs.

For Indonesian stance classification, Setiawan \textit{et al.} \cite{setiawan2024indonesian} proposes a hybrid BiLSTM-Transformer model using indoBERT for word embeddings. BERT-generated vectors are fed into BiLSTM layers to capture contextual information in both directions. The system incorporates data from news and social media, along with preprocessing and feature extraction steps, enabling effective handling of Indonesian linguistic nuances.

In Chinese stance detection, Pu \textit{et al.} \cite{Pu_2024} combines RoBERTa’s semantic strength with BiLSTM’s temporal modelling, using BertTokenizer for proper text handling. BERT-SLEK \cite{Yin_2024} advances this by encoding comments, targets, events, and stances into BERT, incorporating ChatGPT-based event summarization to enhance contextual reasoning and outperform baselines. Zhu \textit{et al.} \cite{Zhu_2023} integrate stance data and non-consecutive semantic features using StanceBERTa, a Cross-modal Transformer, and a Weighted Graph Attention Network (W-GAT), followed by MLP classification. Similarly, Sun \textit{et al.} \cite{Sun_2021} embed domain knowledge from CN-DBpedia via Knowledge Query and Injection, using a visible matrix to reduce noise and boost microblog stance detection.
Alghaslan \textit{et al.} \cite{Alghaslan_2024} fine-tune LLMs—ChatGPT-3.5-turbo, Meta-Llama-3-8B-Instruct, and Falcon7B-Instruct—on the multi-dialectal Arabic MAWQIF dataset for target-specific stance detection. Prompts guide the models to identify the writer’s stance, showcasing LLMs’ effectiveness in Arabic stance tasks. Galal \textit{et al.} \cite{Galal_2024} employ sentence transformers to extract embeddings from MAWQIF tweets using frozen BERT-based PLMs, followed by logistic regression for classification. Both [CLS] and averaged embeddings are used. They also adapt the Parallel-Sum (P-SUM) architecture originally for sentiment analysis to multi-task learning for stance and sentiment detection, enhancing Arabic stance classification performance \cite{Shukla_2024, Al_Hariri_2024, Charfi_2024}.

Cross-lingual approaches offer strong generalization for stance detection. Zhang \textit{et al.} \cite{Zhang_2023} combine a prompt-tuned mBERT (cross-lingual teacher) and a graph-based cross-target teacher to distil knowledge into a student model for zero-shot detection. Similarly, ATOM \cite{10297287} learns language-invariant representations via multilingual models, memory refinement, and adversarial training with Wasserstein-guided language discrimination, using datasets such as X-Stance and R-ita. \cite{schafer2023queen} system fine-tunes mBERT on the CoFE dataset, covering 24 EU languages plus Catalan and Esperanto. It employs sentence-pair classification with segment embeddings, experimenting with both standard and BiLSTM-enhanced architectures on original and translated data.

These studies highlight the growing power of transformer-based models from specialized monolingual models to multilingual giants. These models enable robust stance detection across diverse languages and settings. Methods such as fine-tuning, hybrid BiLSTM-Transformer models, and domain knowledge injection enhance the contextual understanding and performance in challenging scenarios, achieving significant progress in cross-lingual, zero-shot, and low-resource tasks. Notable efforts further cement the transformative potential of these models in stance detection across multiple domains and languages.

\subsection{Hybrid Approaches}
Hybrid approaches have gained attention across various domains for their ability to combine complementary data sources and methodologies, leading to improved performance and broader applicability.
In the context of stance detection, it has gained significant attention in recent years, particularly with the integration of multimodal inputs and external knowledge to enhance performance across domains and languages.
Chunling  \textit{et al.}  \cite{chunling-etal-2023-adversarial} incorporates external knowledge bases through adversarial learning, enabling the model to generalize better across unseen topics. Similarly, Cavelheiro  \textit{et al.} \cite{cavalheiro-etal-2023-stance} explores the combination of textual, demographic, and network-based features within BERT-based models, demonstrating that social context and user metadata can significantly enrich stance prediction accuracy. Bai  \textit{et al.} 
\cite{Bai_2024} fuses image and text modalities using contrastive learning, showing how visual cues from tweets can support nuanced stance understanding, particularly in domains with high visual engagement such as climate discourse. Zhang  \textit{et al.}  \cite{ESC} pushes this further by incorporating entity-level information from both images and text, creating a synergy between factual validation and stance detection in multimodal settings.
Meanwhile, Pu  \textit{et al.}  \cite{pu} introduces emotion-aware multi-task learning where emotional cues extracted from text are fused with stance signals to enhance detection, especially for content related to security or misinformation. These hybrid approaches collectively demonstrate that integrating additional modalities such as images, emotions, entities, and knowledge sources significantly enhances stance classification beyond traditional text-only models.  
Key innovations include cross-modal fusion mechanisms, adversarial and contrastive learning for domain adaptation, and multi-task frameworks that jointly model stance with auxiliary features like emotion or entity relations.  Future work should prioritize:
Unified fusion architectures that balance model complexity with interpretability,
Automated knowledge integration to reduce reliance on manual feature engineering, and
Lightweight hybrid designs for scalable processing of streaming multimodal data.
By addressing these challenges, hybrid stance detection can evolve into a versatile tool for analyzing complex discourse across languages, domains, and evolving sociopolitical contexts.

\section{Target-based Approaches}\label{Section:ta}
In stance detection, the relationship between the model and the target of interest plays a critical role in determining the performance and generalization of the model. This section explores three primary target-based approaches such as in-target, cross-target, and multi-target approaches.
\subsection{In-target Approach}
In-target stance detection refers to models trained and evaluated on the same target or set of targets. This approach assumes prior knowledge of the target and leverages this to enhance model performance. Since both training and testing data are aligned in terms of the target context, models can capture specific linguistic and semantic cues associated with the target, leading to improved performance.
Jiang \textit{et al.} \cite{Jiang_2022} presents a method for stance detection by distilling target-aware prompts. The approach enhances the model's generalization by incorporating prompts from limited labelled data. 
Garg \textit{et al.} \cite{garg-caragea-2024-stanceformer} introduces Stanceformer, a transformer-based architecture designed to enhance attention mechanisms towards specific targets. By incorporating a Target Awareness Matrix, their model boosts self-attention scores associated with the target, resulting in improved stance prediction accuracy during both training and inference.
Li \textit{et al.} \cite{li-etal-2023-new}  move beyond predefined target lists by introducing the task of Target-Stance Extraction (TSE). Their two-stage framework first identifies relevant targets within the input text and then determines the stance expressed towards each predicted target. This approach reflects real-world applications where targets may not be explicitly mentioned.
Ma \textit{et al.} \cite{10.1007/978-981-97-9443-0_7} proposes a novel prompting technique called Chain of Stance (CoS). CoS decomposes stance detection into a sequence of intermediate, stance-related assertions. Leveraging the capabilities of large language models (LLMs), this chain-based reasoning framework significantly boosts classification performance, especially in few-shot learning setups.
Dai \textit{et al.} \cite{DAI2025106956} proposes a hybrid approach combining Logic Tensor Networks (LTNs) with Large Language Models (LLMs) to improve stance classification. It leverages the semantic power of LLMs and the logical reasoning capability of LTNs to incorporate symbolic knowledge into the learning process. 
Ding \textit{et al.} \cite{DING2025128849} explores stance detection using both target-specific and target-invariant approaches. The target-specific approach focuses on fine-tuning models to detect stances in relation to specific targets, while the target-invariant approach leverages general representations for stance detection without focusing on the target. The inclusion of external knowledge helps improve zero-shot performance by enabling the model to generalize stance detection across different, unseen targets.
By aligning training and evaluation on the same targets, these approaches capture domain-specific linguistic patterns and contextual cues more effectively than general-purpose models. Recent advancements highlight several key trends: (1) enhanced target awareness through specialized attention mechanisms and prompt engineering, (2) hybrid architectures combining neural and symbolic reasoning for improved interpretability, and (3) flexible frameworks capable of handling both explicit and implicit target references. While these methods excel in controlled settings, future work should address their adaptability to evolving targets and scalability across languages. The integration of dynamic knowledge updating mechanisms and lightweight adaptation techniques could further bridge the gap between specialized in-target performance and broader zero-shot generalization needs.
 
\subsection{Cross-target Approach}
Cross-target stance detection aims to generalize stance prediction from one target to another, often unseen or unrelated. This requires models capable of identifying stance-relevant signals regardless of the subject. Traditional supervised approaches often struggle in such settings due to their reliance on target-specific patterns. In contrast, LLMs excel by encoding rich semantic representations and transferring knowledge across domains. Their ability to capture deep contextual signals has been further enhanced through domain-adaptive fine-tuning, contrastive learning, and prompt engineering. These capabilities enable robust performance in low-resource and dynamic environments where labelled data is scarce. As public discourse evolves rapidly, scalable stance detection becomes essential. LLMs and transformers address this need by eliminating the dependence on target-specific supervision, forming the backbone of modern, generalizable stance detection systems.

Recent advances in stance detection, especially in cross-target stance detection (CTSD), emphasize contextual knowledge integration, explainable prompting, and meta-learning for better generalization. A key contribution is the LLM-Driven Knowledge Injection framework \cite{Zhang_2024}, which prompts LLMs to extract stance-relevant keywords, rhetorical cues, and implied emotions from text-target pairs. This knowledge is injected into BART, reframing stance detection as a denoising sequence generation task. To further boost semantic alignment, a prototypical contrastive learning scheme projects stance embeddings into a low-dimensional space and aligns them with class-wise prototypes, enhancing intra-class cohesion and inter-class separation.

This line of work aligns with prompt-based frameworks such as PSDCOT \cite{Ding_2024} and MPPT \cite{Ding_COT_2024}, which enhance pre-trained language models with Chain-of-Thought (CoT) reasoning. PSDCOT employs a two-stage process: instruction-based Q\&A for rationale generation, followed by a multi-prompt network that creates stance vectors from diverse semantic perspectives, refined via attention over extracted knowledge. MPPT extends this by aggregating multi-view natural language explanations, incorporating external affective knowledge from sources such as SenticNet. Both approaches highlight the importance of interpretable prompting, semantic alignment, and knowledge fusion for robust cross-domain stance detection.

Complementing these prompting-based approaches, Ji \textit{et al.} \cite{Ji_2022} address CTSD from a meta-learning perspective by enhancing Model-Agnostic Meta-Learning (MAML) with a curriculum-based learning schedule. Their strategy employs inter-target similarity analysis to create an easy-to-hard task sampling curriculum, improving the model's adaptability to new targets by refining optimization paths and data preparation strategies.

Collectively, these methodologies highlight a broader paradigm shift in stance detection: from optimisation-driven transfer methods toward knowledge-rich, interpretable prompting frameworks. The integration of LLM-extracted contextual insights, prototype-based contrastive regularization, and CoT-enhanced semantic reasoning all aim to produce stance models that are not only more accurate across diverse targets but also more explainable and cognitively aligned with human-such as reasoning.

\subsection{Multi-target Approach}
A multi-target approach in stance detection refers to a method that is designed to detect and analyze stances toward multiple targets within the same text. Instead of focusing on a single target, as in traditional stance detection, a multi-target approach is capable of handling texts that express stances towards multiple distinct targets within one sentence or document.
Li \textit{et al.} \cite{li-etal-2021-improving-stance} proposes a novel framework that enhances stance detection by training models on multiple datasets from different domains. This multi-dataset approach enables the model to learn more universal representations, improving its generalization capabilities. Additionally, the authors introduce an Adaptive Knowledge Distillation (AKD) method, which applies instance-specific temperature scaling to teacher and student predictions, further boosting performance and surpassing state-of-the-art results.
Xi \textit{et al.} \cite{9750451} leverages Pretrained Transformer Models (PTMs) and multi-task learning to tackle multi-target stance detection. The model improves performance by simultaneously learning to detect stances towards multiple targets while utilizing task-specific knowledge from related tasks, enhancing generalization. By integrating PTMs with multi-task learning, the approach effectively captures complex relationships between targets and their associated stances across diverse contexts.
Sun \textit{et al.} \cite{10.1145/3544490} introduces a framework that enhances stance detection by integrating multi-target data through an adversarial attention mechanism. This approach identifies and connects topic and sentiment information across posts, facilitating the detection of target-invariant features essential for stance classification. By leveraging adversarial training, the model effectively captures nuanced relationships between topics and sentiments, improving stance detection performance across diverse targets.
Steel \textit{et al.} \cite{steel-ruths-2024-multi} introduces a comprehensive workflow for assessing user opinions on multiple topics simultaneously using Reddit data. It highlights the complexities of multi-target stance detection compared to single-target approaches and demonstrates the effectiveness of their method by replicating known opinion polling results and enabling detailed studies of opinion dynamics over time and across topics.
Avila \textit{et al.} \cite{Avila_2024} explores strategies to enhance stance detection across multiple targets in a multilingual setting. The authors experimented with back-translation to augment training data and label propagation to utilize unlabelled data, aiming to improve model performance.
In conclusion, the effectiveness of the multi-target approach varies based on the context and dataset. For models working across multiple domains, using multi-dataset learning with adaptive knowledge distillation enhances generalization and performance. In multilingual settings, strategies such as data augmentation through back-translation and label propagation are beneficial when labelled data is scarce. Meanwhile, approaches such as multi-task learning and adversarial attention mechanisms excel when capturing complex relationships between targets and stances within a single language or domain, emphasizing the need for task-specific knowledge and nuanced feature extraction. The choice of approach should depend on the complexity of the targets and the diversity of the data.
\section{Agentic and Tool-Augmented LLMs}

Recent advances in large language models have moved stance detection beyond simple prompt-based predictions to more structured, agentic reasoning systems. Earlier LLM approaches usually depend on carefully written prompts or single-step chain-of-thought reasoning to predict stance. These methods are often sensitive to how the prompt is written and struggle with long texts, implicit opinions, and missing background context. Agentic LLMs overcome these limitations by breaking stance detection into multiple steps, retrieving external information when needed, and combining intermediate reasoning results before making the final stance decision.

Lee \textit{et al.} \cite{lee-etal-2025-journalism} introduce Journalism-Guided Agentic In-Context Learning. The model first predicts stance for journalism-motivated components, including headlines, leads, quotations, and conclusions, and then aggregates these predictions to infer the overall article stance.
Zhu \textit{et al.} \cite{zhu-etal-2025-ratsd} propose RATSD, a framework for truthfulness stance detection on social media posts toward factual claims. RATSD retrieves relevant background information, uses LLMs to generate stance-aware analyses, and performs downstream classification.
Retrieval-augmented reasoning has also been employed to support stance-aware factual assessment and explainability. Upadhyay \textit{et al.} \cite{Upadhyay2025} propose a retrieval-augmented framework for health information analysis.
Stance detection is used to check factual accuracy by comparing generated text with retrieved evidence.
In this approach, stance acts as an intermediate reasoning step rather than a final prediction.

Overall, recent work demonstrates a clear transition from single-pass stance classification toward retrieval-augmented and agentic frameworks. Designing stance detection systems with adaptive planning, iterative self-correction, and dynamic tool selection remains an open research direction.

\section{Datasets and Benchmarks}\label{Section:dataset}
This section provides an overview of the datasets and benchmarks commonly used in the field of stance detection. Important benchmarks that are used to assess and contrast the performance of different models in the field of stance detection are also highlighted in this section.
Details of the several datasets utilized for the stance identification task are given in Table ~\ref{tab:datasets}.

\subsection{Unimodal Datasets}
Unimodal datasets used in stance detection primarily consist of text-based data sourced from various platforms such as Twitter, Reddit, news articles, and blogs. These datasets form the foundation of most stance detection tasks due to the widespread availability and ease of collection of textual data.
Prominent among them is the P-STANCE dataset, comprising over 21,000 English tweets annotated for stance, and the COVID-19 vaccine stance dataset, which offers a large-scale corpus with more than two million labelled tweets discussing COVID-19 vaccines. Datasets such as CHeeSE, WT-WT, and X-Stance provide valuable benchmarks for evaluating stance detection systems in political, health, and multilingual contexts. Moreover, resources such as ClimaConvo and ClimateMiSt address emerging domains such as climate misinformation, reflecting the dynamic nature of public discourse.
While many stance detection datasets are based on textual content, such as tweets, news articles, and forum discussions, datasets like THUMOS14 and ActivityNet-1.3 represent a different modality—video.
In contrast, text-based datasets dominate the stance detection landscape, offering the advantage of rich linguistic features that are amenable to deep natural language processing (NLP). This focus on the textual modality allows for detailed syntactic, semantic, and discourse-level analysis, facilitating the development of robust NLP-based stance classification models.
However, this heavy reliance on textual datasets reveals a significant gap. Given the increasing consumption and production of video content across platforms such as YouTube, TikTok, and Instagram, extending stance detection to other modality based contexts is both timely and necessary.

\subsection{Multilingual Datasets}
Multilingual stance detection is gaining prominence due to the global nature of social discourse.
Datasets such as HIPE, X-Stance, and Catalonia Independence Corpus are designed to evaluate cross-lingual transfer and multilingual understanding. 
These corpora include samples in languages such as German (DE), French (FR), Italian (IT), Catalan (CA), Arabic (AR), Hindi (HN), and Chinese (ZH), broadening the applicability of stance detection models beyond English. For instance, the VaxxStance dataset provides data in European and Spanish languages, while MAWQIF and MARASTA target Arabic-language social media content, addressing low-resource challenges in Arabic NLP. The presence of these multilingual datasets enables researchers to test the robustness of models across languages, paving the way for more inclusive and globally relevant NLP systems.

Although English remains the dominant language used in online platforms, focusing on low-resource languages in stance detection is critical and important. Such efforts can contribute to preserving endangered languages and cultural narratives, ensuring that voices from marginalized communities are represented in digital discourse. In this way, multilingual stance detection research serves both technological advancement and the cultural imperative of sustaining linguistic diversity in the age of global communication.

\subsection{Multimodal Datasets}
Multimodal stance detection has additional modalities such as images ,audio, and  videos to complement text, reflecting on the multi-format nature of social media communication. Datasets such as CMFC, MmMtCSD, and Fakeddit include both textual and visual information, enabling models to better understand context, sarcasm, and nuanced opinions. 

 On the other hand, the Pushshift dataset,  a text-based dataset, enhances multimodal research with metadata or other information. Such datasets are crucial for developing models that go beyond linguistic cues to incorporate visual semantics, which is especially important in understanding complex or ambiguous stances conveyed through memes, infographics, or video content.

 The ability to contribute to the construction of multimodal datasets is increasing with the development of technologies such as OCR (Optical Character Recognition) and generative models such as GPT-4o. Datasets can be enhanced with semantic annotations by automatically generating written explanations of visual content, picture captioning, or video summarization.  Additionally, particularly for low-resource or underrepresented content forms, data augmentation techniques such as generating synthetic images \cite{shah-etal-2024-memeclip} can help overcome data shortages and enhance model robustness.

\begin{table*}[!ht]
  \centering
  \resizebox{\linewidth}{!}{  
  \begin{tabular}{|l|l|l|l|l|l|l|l|}
    \hline
    \textbf{Dataset} & \textbf{Author} & \textbf{Year} & \textbf{Source} & \textbf{Available at} & \textbf{Language} & \textbf{\# Samples} & \textbf{Modality}\\
    \hline
    HIPE & M. Ehrmann \textit{et al.}\cite{ehrmann2020introducing} & 2020 & News Articles & \href{https://github.com/impresso/CLEF-HIPE-2020/tree/master/data}{Link} & DE, FR, EN &  14,555 & Text\\
    Czech & Hercig \textit{et al.}\cite{SLON2017-Stance} & 2017 & News
server & -& CS & 1,455 & Text\\
    NEWSEYE & Hamdi \textit{et al.} \cite{10.1145/3404835.3463255} & 2021 & News Articles &- & FR, DE, Finnish, Swedish & 30,580 & Text\\
    P-STANCE & Li \textit{et al.} \cite{Li_2021}& 2021 & Twitter & \href{https://drive.google.com/drive/folders/1so8lY1XKpnhUtTvb15edEz6aeHt7CSuh}{Link}& EN & 21,574 & Text\\
    rumorEval & Gorrell, \textit{et al.}\cite{Gorrell_2019} & 2019 & Twitter & \href{https://figshare.com/articles/dataset/rumorEval_2019_data/8845580?file=16188500}{Link}& EN & 8,574 & Text\\
    VaxxStance &  Agerri \textit{et al.} \cite{agerri2021vaxxstance} & 2021 & Twitter  & \href{https://vaxxstance.github.io/}{Link} & EU, ES &4,081 & Text, Metadata\\
    COVID-19 VACCINE STANCE & Cotfas \textit{et al.} \cite{Cotfas_2021} & 2021 & Twitter & -& EN & 2,349,659 & Text\\
     CHeeSE & Mascarell \textit{et al.} \cite{Mascarell_2021}& 2021 & News Articles & \href{https://github.com/MTC-ETH/CHeeSE}{Link}& DE & 3,693 & Text\\
    COVID-19-Stance & Glandt \textit{et al.} \cite{Glandt_2021} & 2021 & Twitter & \href{https://github.com/kglandt/stance-detection-in-covid-19-tweets}{Link}& EN& 36,762 & Text\\
     Catalonia Independence Corpus & Zotova \textit{et al.} \cite{zotova2020multilingual} & 2021 & Twitter & \href{https://github.com/ixa-ehu/catalonia-independence-corpus}{Link}& CA, ES & 46,099  & Text\\
     DeInStance & Gohring \textit{et al.} \cite{gohring2021d} & 2021 & Smartvote & \href{https://github.com/ZurichNLP/xstance}{Link} & DE & 1,000 & Text\\
     MAWQIF & Alturayeif \textit{et al.} \cite{Alturayeif_2022} & 2022 & Twitter & \href{https://github.com/NoraAlt/Mawqif-Arabic-Stance}{Link} & AR & 4,121 & Text\\
     MARASTA & Charfi \textit{et al.} \cite{charfi-etal-2024-marasta} & 2024 & Twitter, YouTube & - & AR & 4,657 & Text\\
     WT-WT & Conforti \textit{et al.} \cite{Conforti_2020} & 2020 & Twitter & \href{https://github.com/cambridge-wtwt/acl2020-wtwt-tweets}{Link} & EN & 51,284 & Text \\
     X-Stance & Vamvas \textit{et al.}\cite{vamvas2020xstance} & 2020 & Smartvote & \href{https://github.com/ZurichNLP/xstance}{Link} & DE, FR, IT & 67,271 & Text\\
     VAST & Allaway \textit{et al.} \cite{Allaway_2020} & 2020 & ARC corpus & \href{https://github.com/emilyallaway/zero-shot-stance}{Link} &EN & 18,545 & \\
     StanceUS, StanceIN & Dutta \textit{et al.} \cite{Dutta_2022} & 2022 & Twitter & \href{https://drive.google.com/file/d/1kJuNjSGwT3riZFyMsvm28TBbjYY8neER/view}{Link} &EN & 8,007 & \\
     ClimaConvo & Shiwakoti \textit{et al.} \cite{shiwakoti-etal-2024-analyzing} & 2024 & Twitter & \href{https://github.com/shucoll/ClimaConvo}{Link} & EN & 15,309 & Text\\
     R-ITA, E-FRA & Lai \textit{et all.} \cite{Lai_2020}& 2020 & Twitter & &  IT, FR & 833 & Text\\
     AuSTR & Haouari \textit{et al.} \cite{Haouari_2023} & 2023 & Twitter & \href{https://github.com/Fatima-Haouari/AuSTR}{Link}& AR & 811 & Text\\
      STANDER & Conforti \textit{et al.} \cite{Conforti_STANDER_2020} & 2020 & Newspapers, Magazines & \href{https://github.com/cambridge-wtwt/emnlp2020-stander-news}{Link}& EN & 3,291 & Text\\
      EZ-STANCE & Zhao \textit{et al.} \cite{Zhao_EZ_2024} & 2024 & Twitter & \href{https://github.com/chenyez/EZ-STANCE}{Link} & EN & 47,316 & Text\\
      ClimateMiSt  & Choi \textit{et al.} \cite{Choi_2025} & 2025 & Twitter & -  & EN & 2,008 & Text\\
      CTSDT & Li \textit{et al.} \cite{Li_2023} & 2023 & Twitter & - & EN & 53,861 & Text\\
      AllSides & Li \textit{et al.} \cite{Li_2019} & 2019 & Online News Articles, Twitter & - & EN & 10,385 &  Text\\
      ExaASC & Jaziriyan \textit{et al.} \cite{Jaziriyan_2021} & 2021 & Twitter & \href{https://github.com/exaco/exaasc}{Link}& AR & 9,566 & Text\\
      StEduCov & Hamad \textit{et al.} \cite{Hamad_2022} & 2022 & Twitter & \href{https://ieee-dataport.org/9221}{Link}& EN & 16,572 & Text\\
   
    COVIDLies & Hossain \textit{et al.} \cite{Hossain_2020} & 2020 & Twitter & \href{https://ucinlp.github.io/covid19}{Link}& EN & 6761 & Text\\
      MSDD & Hu \textit{et al.} \cite{Hu_2023} & 2023 &TV shows & -& EN &1,296 &Text, Images\\
      CMFC & Zhang \textit{et al.} \cite{Zhang_ESCNet_2024} & 2024 & News Websites &- & ZH & 26,000 & Text, Image\\
      MmMtCSD & Niu \textit{et al.} \cite{10.1145/3664647.3681416} & 2024 & Reddit & \href{https://github.com/nfq729/MmMtCSD}{Link} &EN & 21,340 & Text, Image\\
      THUMOSIS14 & Jiang \textit{et al.}  \cite{THUMOS14}  & 2014 &YouTube  & \href{https://github.com/open-mmlab/mmaction2/blob/main/tools/data/thumos14/README.md}{Link} & -& 2,064 & Videos\\
      ActivityNet-1.3 &  Heilbron \textit{et al.} \cite{caba2015activitynet} & 2015 &
      Youtube & \href{http://activity-net.org/}{Link}& -& 19,994 & Videos\\
      CSD & Li \textit{et al.} \cite{Li_Convothreads_2023} & 2023 & Social Media Sites & \href{https://github.com/nfq729/MT-CSD}{Link} &YUE & 500 Posts, 5,376 Comments & Text, Metadata\\
      STANCEOSAURUS & Zheng \textit{et al.} \cite{Zheng_2022} & 2022 & Twitter & \href{https://tinyurl.com/stanceosaurus}{Link} & EN, HN, AR & 28,033 & Text \\
      AraStance & Alhindi \textit{et al.} \cite{alhindi-etal-2021-arastance} & 2021&Online Websites &\href{https://github.com/Tariq60/arastance}{Link} & AR & 4,063  & Text \\
      ANS &  Khouja\textit{et al.} \cite{khouja-2020-stance} & 2020 &News Article  & \href{https://github.com/latynt/ans}{Link} & AR & 3,786 & Text \\
      ArCOV19-Rumors &  Haouari\textit{et al.} \cite{haouari-etal-2021-arcov19} &2021  & Online Websites & \href{https://github.com/TianBian95/BiGCN}{Link} & AR & 9,413 & Text \\
      GenderStance &  Li \textit{et al.} \cite{li-zhang-2024-pro} & 2024 & Online Websites & \href{https://github.com/chuchun8/GenderStance}{Link} & EN & 36,000 & Text \\
      ORCHID & Zhao and Wang\textit{et al.} \cite{zhao-etal-2023-orchid} & 2023 & Online Debate Videos & \href{https://github.com/xiutian/OrChiD}{Link}& ZH & 436 summaries, 14,133 utterances & Text\\
      C-MTCSD & Niu and Yang\textit{et al.} \cite{10.1145/3701716.3715307} & 2025 &Weibo & \href{https://github.com/yangyi626/C-MTCSd}{Link}& ZH &  24,264 & Text\\
      UstanceBR & Pereira \textit{et al.} \cite{Pereira_2024} & 2026 & Twitter & - & PT & 46,800 & Text\\
      TSD-CT& Zhu\textit{et al.} \cite{10.1145/3746252.3761622} & 2025 &Twitter & \href{https://github.com/idirlab/stancedatacollection}{Link}& EN &  5,331 & Text\\
     PoliStance-TR& Unal\textit{et al.} \cite{unal-etal-2025-polistance} & 2025 &X & \href{https://anonymous.4open.science/r/polistance-tr-22C4/}{Link}& TR &  7,923 & Text\\
      \hline
  \end{tabular}
  }
  \caption{\label{tab:datasets}
    Datasets for Stance Detection
  }
\end{table*}

\subsection{Evaluation Metrics}
Accuracy, Precision, Recall, and F1-score (F1-Macro) are assessment measures used to evaluate model performance in stance detection. The accuracy is defined as:

\begin{equation}
\text{Accuracy} = \frac{\text{Correct Predictions}}{\text{Total Predictions}}
\label{eq:accuracy}
\end{equation}
Precision, recall, and F1-score for each class are defined as follows:

\begin{equation}
\text{Precision}_i = \frac{\text{True Positives}_i}{\text{True Positives}_i + \text{False Positives}_i}
\end{equation}
\begin{equation}
\text{Recall}_i = \frac{\text{True Positives}_i}{\text{True Positives}_i + \text{False Negatives}_i}
\end{equation}
\begin{equation}
\text{F1}_i = 2 \times \frac{\text{Precision}_i \times \text{Recall}_i}{\text{Precision}_i + \text{Recall}_i}
\end{equation}
Where \(i\) refers to a specific class (favor, against, or neutral). The macro average for these metrics is computed by averaging the scores across all classes:

\begin{equation}
\text{Macro Precision} = \frac{1}{N} \sum_{i=1}^{N} \text{Precision}_i
\end{equation}
\begin{equation}
\text{Macro Recall} = \frac{1}{N} \sum_{i=1}^{N} \text{Recall}_i
\end{equation}
\begin{equation}
\text{Macro F1} = \frac{1}{N} \sum_{i=1}^{N} \text{F1}_i
\end{equation}

Where \(N\) is the total number of classes. These metrics ensure that the model's performance is evaluated in a balanced manner, especially when dealing with imbalanced classes.
For a balanced performance, the macro average of the metrics is used primarily, as it treats each class equally, regardless of any class imbalance. 
\subsection{Results}

\begin{table}[h!]
\centering
\caption{F1-Macro scores of different methods across datasets.}
\label{tab:f1_macro_results}
\resizebox{\textwidth}{!}{
\begin{tabular}{|c|c|c|c|c|}
\hline &
\textbf{Dataset}&\textbf{Method}& \textbf{Author}  &   \textbf{F1-Macro (\%)} \\
\hline
\multirow{3}{*}{
\makecell{Encoder \\ -based LLMs}} &CHeeSE & German BERT &Mascarell \textit{et al.} \cite{Mascarell_2021}  &  58.40 \\
&COVID-19-Stance &   Wikipedia + RoBERTa&Yan \textit{et al.} \cite{10654680}  & 85.76 \\
&\makecell{Catalonia \\Independence} &   mBERT&Zotova \textit{et al.} \cite{ZOTOVA2021114547}  & 85.76 \\
&DeInStance &   ${\text{BERT}}_{ft}$ &Gohring \textit{et al.} \cite{gohring2021d}  & 70.00 \\
&Mawqif &  Multi-Task Learning &Alturayeif \textit{et al.} \cite{Alturayeif2024}  & 85.10 \\
&WT-WT & Contrastive Learning & Jiang \textit{et al.} \cite{JIANG2023103361}  & 80.90 \\
&X-Stance & ${\text{XLM-R}}_{ft}$ &  Barriere \textit{et al.} \cite{vamvas2020xstance}  & 74.62 \\
&VAST & Wikipedia + RoBERTa &  Yan \textit{et al.} \cite{10654680}  & 80.70 \\

&R-ITA, E-FRA & Label-Encoding &  Hardalov \textit{et al.} \cite{Hardalov_2022}  & 73.00 \\
&AuSTR & $\text{BERT}_{ft}$&  Haouari \textit{et al.} \cite{Haouari2024}  & 83.20 \\
&CTSDT & BERT & Li \textit{et al.} \cite{Li_2023} & 62.90 \\
&CMFC & Chinese BERT+ResNet-50 & Zhang \textit{et al.} \cite{Zhang_ESCNet_2024} & 85.40 \\
&STANCEOSAURUS & BERTweet & 
Zheng \textit{et al.} \cite{10.1145/3664647.3681416}& 51.75 \\
&AraStance  & ARBERT & 
Alhindi \textit{et al.} \cite{alhindi-etal-2021-arastance} & 78.00 \\
&ANS & BERT & 
Khouja \textit{et al.} \cite{khouja-2020-stance} & 76.70 \\
&ArCOV19-Rumors & $\text{BERT}_{ft}$ & Haouari \textit{et al.} \cite{Haouari2024} & 74.50 \\
&GenderStance & BART-MNLI & 
Li \textit{et al.} \cite{li-zhang-2024-pro} & 64.30 \\
&MARASTA & QARiB & Bensalem \textit{et al.} \cite{Bensalem_2025}  & 80.90 \\
&CLiCS & BERTweet+Sentiment&Upadhyaaya \textit{et al.} \cite{Upadhyaya_2023} & 86.58 \\
&rumorEval-2019 & BERT & Li \textit{et al.} \cite{zhu-etal-2025-ratsd} & 76.00 \\
&tWT-WT & BERT & 
Kaushal \textit{et al.} \cite{kaushal-etal-2021-twt} & 36.50 \\
&Debating Europe & XLM-R & 
Barriere \textit{et al.} \cite{barriere-etal-2022-debating} & 71.40 \\
&DEBAGREEMENT & RoBERTa & Lou \textit{et al.} \cite{akash2024can} & 66.91 \\
&ClimaConvo & RoBERTa &  Montesinos \textit{et al.} \cite{reyes-montesinos-rodrigo-2024-hamison}  & 74.95 \\
&STANDER & \makecell{BERT + Finegrained \\ Evidence Retrieval} & Conforti \textit{et al.} \cite{Conforti_STANDER_2020} & 45.70 \\
&CSD &  
BERT+CNN & Akash \textit{et al.} \cite{akash2024can} & 79.00 \\
&PoliStance-TR& ELECTRA& Unal\textit{et al.} \cite{unal-etal-2025-polistance}&0.82\\

\hline
\multirow{3}{*}{\makecell{Decoder \\ -based LLMs}} 
&P-Stance  & LLM Prompting&Li \textit{et al.} \cite{app15094612} & 86.52 \\
&VaxxStance & GPT-4&Zhang \textit{et al.} \cite{agerri2021vaxxstance}  &  53.10 \\
&EZ-STANCE & Mistral & Akash \textit{et al.} \cite{akash2024can} & 79.00 \\
&ClimateMiSt & GPT-4 & Choi \textit{et al.} \cite{Choi_2025} & 89.53 \\

&COVIDLies & Zephyr & Zhu \textit{et al.} \cite{zhu-etal-2025-ratsd} &  67.87 \\

&MmMtCSD& LLaMA2-7b+GPT4-Vision & 
Niu \textit{et al.} \cite{10.1145/3664647.3681416} & 71.85 \\

&ORCHID & GPT-3.5 & 
 Zhao \textit{et al.} \cite{zhao-etal-2023-orchid} & 88.13 \\
 &C-MTCSD & GPT-4.0 & 
 Niu and Yang \textit{et al.} \cite{10.1145/3701716.3715307} & 64.07 \\
 &TSD-CT & GPT-3.5 & 
 Zhu \textit{et al.} \cite{zhu-etal-2025-ratsd} & 87.13 \\

\hline
\end{tabular}
}
\end{table}
Table \ref{tab:f1_macro_results} presents the F1-Macro scores of various stance detection methods across a wide range of datasets.
The presented table provides a comprehensive comparison of F1-Macro scores achieved by various stance detection methods across multiple datasets, categorizing approaches into encoder-based and decoder-based large language models (LLMs). Encoder-based models, including BERT variants such as German BERT, mBERT, RoBERTa and hybrid architectures such as BERT+CNN, BERTweet+Sentiment), demonstrate strong performance. These models leverage pre-trained contextual embeddings and are typically fine-tuned using supervised learning, with some employing advanced techniques such as multi-task learning, contrastive learning, and multimodal fusion. The robustness of encoder-based models is evident in their widespread application across diverse datasets, including CHeeSE \cite{Mascarell_2021}, COVID-19-Stance \cite{Glandt_2021}, and AraStance \cite{alhindi-etal-2021-arastance}, where they achieve competitive results. However, these models exhibit limitations, including a heavy reliance on labelled data for fine-tuning, which restricts their effectiveness in low-resource languages and domains. Additionally, their performance often degrades when applied to unseen targets, highlighting challenges in cross-domain generalization. Computational demands for training and inference further constrain their scalability.  

In contrast, decoder-based LLMs, such as GPT-4, Mistral, and LLaMA2, represent an emerging paradigm in stance detection, particularly for zero-shot and few-shot scenarios. These models, exemplified by their application in ClimateMiSt \cite{Choi_2025} and ORCHID \cite{zhao-etal-2023-orchid}, leverage generative capabilities and instruction fine-tuning to achieve notable performance, with GPT-4 attaining an F1-Macro score of 89.53\% on ClimateMiSt \cite{Choi_2025}. The flexibility of decoder-based models eliminates the need for extensive fine-tuning, making them suitable for low-resource settings. However, their performance is inconsistent, as seen in the substantial drop to 53.10\% on VaxxStance \cite{agerri2021vaxxstance}, underscoring sensitivity to dataset-specific factors.  

The table further reveals broader challenges in stance detection, including the inherent ambiguity and sarcasm in social media text, which complicate stance classification. Multilingual and dialectal variations, as observed in datasets such as MARASTA \cite{charfi-etal-2024-marasta} and AraStance \cite{alhindi-etal-2021-arastance}, introduce additional complexity, with performance often varying across languages. Despite these challenges, the integration of multimodal data, as demonstrated by CMFC \cite{Zhang_ESCNet_2024} and MmMtCSD \cite{10.1145/3664647.3681416}, offers promising avenues for improving model robustness by incorporating visual and textual context. The advantages of encoder-based models lie in their reliability and established efficacy, while decoder-based models provide adaptability and reduced dependency on labelled data. Future research should prioritize enhancing cross-domain generalization, optimizing computational efficiency, and developing interpretable frameworks to advance the field of stance detection. The table thus serves as a valuable reference for understanding the current landscape, methodological diversity, and persistent limitations in stance detection research.

\section{Applications}\label{Section:application}
\begin{figure*}[!ht]
    \centering
    \captionsetup{justification=justified}   \includegraphics[scale=0.8]{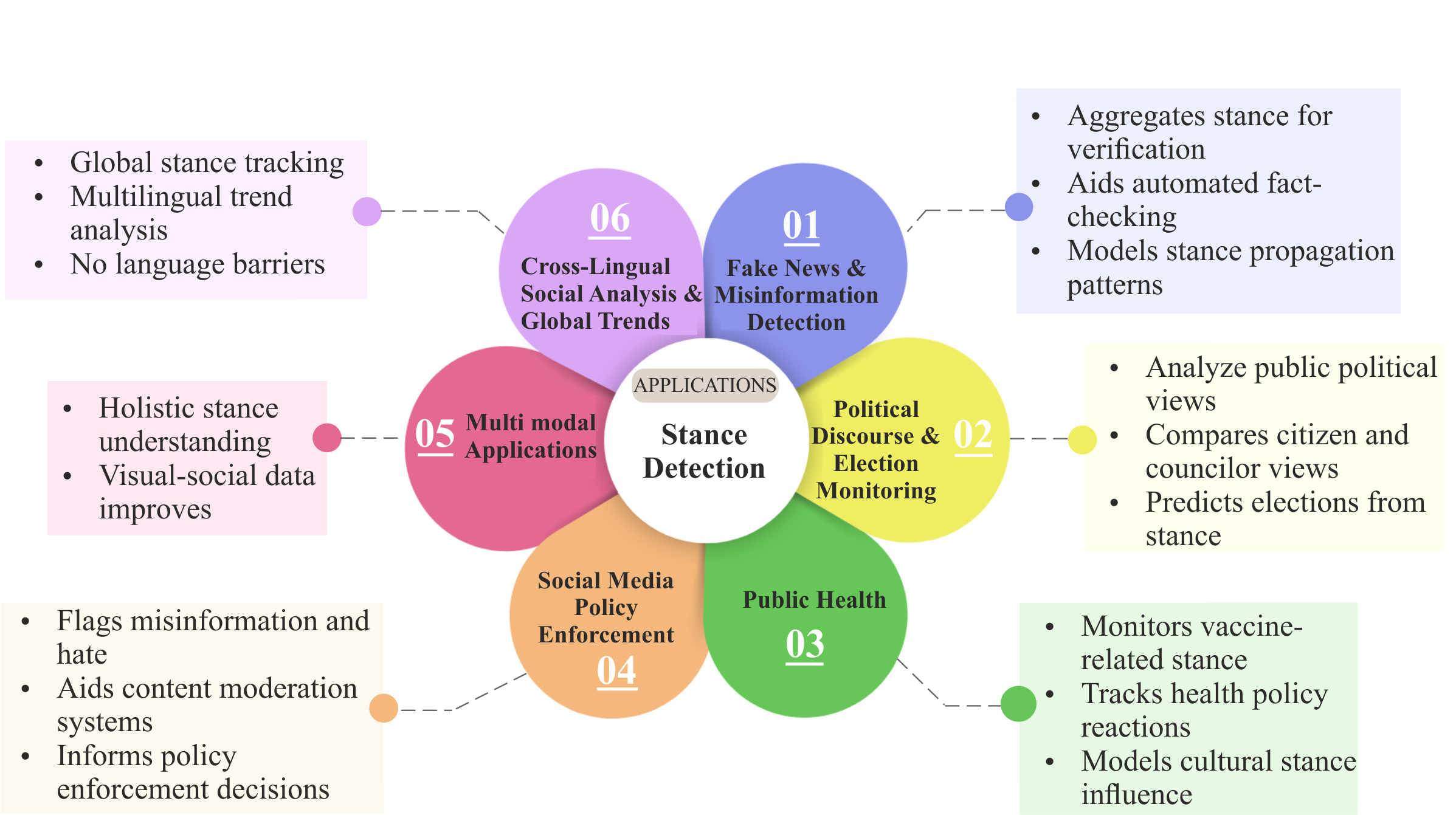}
    \caption{Applications of Stance Detection}
    \label{fig:fig12}
\end{figure*}

This section provides an in-depth exploration of the diverse applications of stance detection across various domains. 
We explore the real-world applications of stance detection in this section. 
One of its key applications is fake news detection, where identifying the stance of the claim helps in assessing its credibility.  
The applications of Stance detection are diverse and span various domains, as illustrated in Figure \ref{fig:fig12},  highlighting the key areas where stance detection techniques are actively applied.

\subsection{Fake News and Misinformation Detection}
Stance detection plays an important role in identifying and combating fake news and misinformation.  It strengthens fact-checking efforts by detecting patterns where a large volume of ``deny'' stances may indicate a potential falsehood.  When stances from numerous users are aggregated, they provide a collective judgment that supports the assessment of information credibility. For example, detecting a significant number of “deny” responses may signal a possible falsehood, aiding in early intervention. This technique enhances traditional fact-checking by revealing patterns in user opinion. In addition, stance detection offers contextual insights that complement content-focused misinformation detectors. This combination of content-based and network-driven features improves the detection of false information by reflecting how users engage with it.
In a thorough analysis, Hardalov \textit{et al.} \cite{hardalov-etal-2022-survey} classify stance detection techniques according to their data sources, model types, and application domains, highlighting their applicability in misinformation detection. 
Sengan \textit{et al.} \cite{10120685} propose a hybrid approach that integrates textual and visual cues with stance information. By combining features from multiple modalities—such as written text, imagery, and user reactions—the system forms a more comprehensive understanding of the content, boosting the detection accuracy. Similarly, a graph-based approach by Soga \textit{et al.} \cite{SOGA202426} to model the relationship between users and post stances. Together, these developments demonstrate how effective stance detection is in combating false information.  Stance detection algorithms provide deeper, context-rich insights that improve the resilience of fake news detection systems by integrating social structure, network behaviour, and multimodal data.

\subsection{Political Discourse and Election Monitoring}
In the political sphere, attitude detection makes it possible to track public sentiment toward political candidates, parties, policies, and election campaigns in real time. Analyzing social media posts, news articles, and forum discussions, it helps to map ideological divides and understand how different communities align or oppose key political narratives. 
The P-STANCE dataset \cite{Li_2021} provides 21,574 tweets labelled for stance toward Donald Trump, Joe Biden, and Bernie Sanders, collected using partisan hashtags to capture a broad range of opinions. Each tweet is annotated as ``Favor'' or ``Against,'' supporting the development of models capable of handling nuanced and polarized content.
Expanding on this, Caceres-Wright \textit{et al.} \cite{Caceres_Wright_2024} introduce the concept of explicit stance detection, curating a dataset of over 1,000 tweets across 13 political figures from the 2020 U.S. election.
Further extending stance detection beyond English, Maia \textit{et al.} \cite{Maia_2022} introduce a Portuguese corpus of political bill comments from the Brazilian Chamber of Deputies website, focusing on cross-target stance detection. 
Building on the need to understand political attitudes at a more local level, Senoo \textit{et al.} \cite{Senoo_2024} compare the stance of citizens and city councilors on political issues by analyzing citizen tweets and city council meeting minutes across three Japanese cities.
Further emphasizing the dynamic nature of stance across related events, Zhang \textit{et al.} \cite{Zhang_Trump_2023} explores connected behavior prediction by modeling how users' stances on topics such as wearing masks and racial equality during COVID-19 could predict their stance toward Donald Trump in the US Election 2020. 
By collecting Twitter data, user profile information, and retweet graphs, and by labeling stance using strong stance-expressing hashtags, the study shows that prior stances are strong predictors of future behavior, achieving high predictive performance.
Expanding on the intersection between social media and political outcomes, Bayrak \textit{et al.} \cite{Bayrak_2023} also explore predicting election results via social media for the 2018 Turkish Presidential Election. 
By detecting stance through retweet-graph-based methods, and applying four different counting methods—simple user counting (SC), city-based weighted counting (CBWC), closest-city-based prediction (CCBP), and using former election results (UFERs)—the study shows how social media data can reduce bias and improve prediction accuracy. Notably, location-based methods, particularly those leveraging sociological similarities between cities (CCBP) and former election results (UFER), significantly outperformed traditional polls and basic methods, suggesting that social media can indeed serve as an alternative medium for conducting political polls.
These findings highlight the importance of location-based information, domain-specific data, and temporal behaviour connections for predicting political stances and outcomes.

\subsection{Public Health}
Stance detection is increasingly valuable in public health communications, particularly in understanding public reactions to health advisories, vaccine drives, and medical research. During events such as the COVID-19 pandemic, detecting stances toward vaccination, mask mandates, and public health policies helped authorities gauge public sentiment and tailor communication strategies. It also aids in identifying clusters of misinformation that could lead to vaccine hesitancy or public health non-compliance.
In the Social Media Mining for Health (SMM4H) 2022 shared task (2024),  Davydova \textit{et al.} \cite{Davydova_2024} introduced a dataset of 9,226 tweets about COVID-19 mandates, with 2,070 manually annotated for vaccine mandates. The study explored early fusion of tweets and claims, dual-view architectures, and syntactic feature incorporation, fine-tuning models such as BART with syntax features, DAN-BERT, and a COVID-19-specific BERT-large for stance prediction, while emotion analysis was conducted using DistilRoBERTa. Results demonstrated the effectiveness of feature aggregation, syntactic enrichment, and transfer learning across claims for health-related stance detection.
Putra \textit{et al.} \cite{Putra_2022} addressed COVID-19 vaccine-related stance detection in Indonesia by combining multi-task aspect-based sentiment analysis (ABSA) with social features. A BiGRU-BERT framework extracted aspect terms through POS tagging and dependency parsing, identified aspect categories and sentiments, and incorporated social attributes such as profile and network features. The multi-task approach significantly outperformed text-only baselines, underscoring the benefit of integrating aspect-level sentiment and community behavior cues.
In multilingual settings, a Spanish corpus of COVID-19 vaccination tweets was developed by Martinez \textit{et al.} \cite{Mart_nez_2023}, comprising 2,801 manually annotated tweets with high inter-annotator agreement. A semi-supervised self-training SVM model expanded the corpus to 11,204 tweets. Sentence-level deep learning embeddings and density-based clustering were utilized for content exploration, demonstrating the scalability and effectiveness of corpus expansion techniques for Spanish-language public health discourse.
Shan \textit{et al.} \cite{Shan_2024} further investigated the cultural determinants of public stances on COVID-19 preventive measures across 95 countries, employing Hofstede’s cultural dimensions. Regression analysis revealed that factors such as power distance, individualism, and uncertainty avoidance systematically influenced stance distributions, with online stances toward face coverings mediating the relationship between cultural profiles and COVID-19 case and death rates. These findings emphasize the importance of culturally sensitive communication strategies and the strategic use of digital platforms to shape public health behaviors.
Additionally, the COVID-19-Stance dataset by Glandt \textit{et al.} \cite{Glandt_2021} provided a targeted resource for analyzing stances toward specific pandemic-related topics in the United States, including stay-at-home orders, mask-wearing, school closures, and public figures. Comprising 6,133 manually annotated tweets, baseline models such as BERT, BERT-NS, and BERT-DAN were evaluated, with self-training and domain adaptation approaches demonstrating improvements by leveraging unlabelled and related-task labelled data.
Together, these studies demonstrate the crucial role of domain adaptation, multi-task learning, cultural analysis, and corpus expansion in advancing stance detection for dynamic and multilingual public health crises.

\subsection{Social Media Policy Enforcement} 
Social media platforms can leverage stance detection to enforce community guidelines and reduce the spread of harmful content. By identifying posts that deny established facts such as climate change denial, election fraud claims, or promote harmful ideologies, stance detection assists in content moderation. 

Recent advancements in stance detection increasingly leverage large language models (LLMs) to overcome data scarcity and improve cross-domain generalization. The Tweets2Stance (T2S) framework by Gambini \textit{et al.} \cite{Gambini_2024} illustrates this trend by using LLMs pre-trained on natural language inference tasks for zero-shot stance detection across a user's X (Twitter) timeline. 
Similarly, Abeysinghe \textit{et al.} \cite{Abeysinghe_2022} use sentence transformers and semantic similarity to detect stance toward web articles in a fully unsupervised, scalable manner.
Complementing the trend toward minimal supervision, the Stance-aware Reinforcement Learning Framework (SRLF) by Yuan \textit{et al.} \cite{Yuan_2021} addresses rumor detection by automatically generating weak stance labels and learning a reinforcement policy to select the most useful ones for improving classification. 
To further address the challenge of limited annotated data, the Stance-aware Multi-Domain Multi-Task (S-MDMT) model by Wang \textit{et al.} \cite{Wang_2021} improves target-specific stance classification through adversarial learning and the use of target descriptors. 
Extending beyond single-instance analysis, Li \textit{et al.} \cite{Li_2023} introduce the novel task of Conversational Stance Detection (CSD), which accounts for the hierarchical and contextual nature of conversation threads. Based on data from six major Hong Kong-based social media platforms, the study constructs a dedicated CSD dataset that preserves full thread structures. To model this context, the proposed Branch-BERT framework combines subbranch-level semantic representations generated via BERT with a TextCNN-based n-gram feature extractor. 

Together, these developments show that with the integration of LLMs, multi-task learning, and contextual modeling, stance detection is becoming a practical and powerful tool for fostering healthier online environments.
\subsection{Multimodal Applications} 
Recent work in stance detection increasingly adopts multimodal approaches to capture the complexities of online discourse. Niu \textit{et al.} \cite{Niu_2024} introduce the MmMtCSD dataset with 21,340 annotated examples involving text, targets (“Tesla” and “Bitcoin”), and images, with 66\% of conversations closely tied to image content. To address challenges of multimodal inference and contextual dependence, the authors propose the MLLM-SD framework, which integrates a textual encoder, Vision Transformer (ViT), and caption encoding, and fine-tunes LLaMA using low-rank adaptation (LoRA). Experiments confirm MLLM-SD’s effectiveness in complex, multi-turn conversations.
Kuo \textit{et al.} \cite{Kuo_2024} explore multimodal stance detection in political discourse, focusing on Facebook fan pages during the 2021 Taiwanese referendums. An unsupervised model integrates text, sentiment, images, posting times, and user interactions. Sentence-BERT (S-BERT) is used for text embeddings, BEiT for visual features, and GraphSAGE builds a graph of posts to group similar stances. Results show that incorporating visual and social features significantly improves stance detection, highlighting the value of multimodal approaches in political analysis.

Khiabani \textit{et al.} \cite{Khiabani_2024} propose the cross-target text-net (CT-TN) model for few-shot stance detection on social media. CT-TN aggregates multimodal embeddings from textual content and network-based features (followers, friends, likes) through three components: text-based classification, network encoding, and output aggregation via majority voting. Using the P-Stance dataset, experiments show CT-TN’s effectiveness with limited training data, and ablation studies highlight the contribution of each module. This work emphasizes the potential of few-shot learning combined with multimodal features for stance detection.
By integrating multiple modalities, these approaches provide a more comprehensive understanding of stance in both social and political discourse, where visual and social cues often play a crucial role in shaping opinions. These advancements suggest that multimodal methods, including the use of graph-based embeddings and few-shot learning, will continue to be instrumental in advancing stance detection, offering a more nuanced and robust analysis of digital conversations and political ideologies.

\subsection{Cross-Lingual Social Analysis and Global Trends}
Cross-lingual stance detection enables monitoring of global events, public health crises, and political movements without language barriers. It supports trend analysis across cultures and regions by identifying commonalities and differences in stances. Such capabilities are critical for multinational organizations, humanitarian efforts, and global media outlets aiming to understand and respond to worldwide sentiment shifts effectively.

Giddaluru \textit{et al.} \cite{Giddaluru_2024} tackle stance detection in Hinglish, a low-resource code-mixed language, using BART-large-MNLI under zero-shot, few-shot, and N-shot settings to generalize across political discourse and diverse topics. Similarly, the VaxxStance dataset by Agerri \textit{et al.} \cite{agerri2021vaxxstance} offers a cross-lingual benchmark on vaccine-related discourse in Basque and Spanish, with over 8 million tweets and rich contextual metadata such as user graphs. It supports multiple evaluation tracks—close, open, and zero-shot—to foster cross-lingual transfer learning.

The Adversarial Topic-Aware Memory Network (ATOM) proposed by Zhang \textit{et al.}  \cite{10297287} tackles zero-resource cross-lingual stance detection by mining topic representations via unsupervised clustering on multilingual embeddings and using an iterative adversarial memory network to extract transferable, topic-aware features. Evaluated on multilingual political datasets, ATOM outperforms existing methods, highlighting the effectiveness of adversarial training and topic-based abstraction.
Zhang \textit{et al.} \cite{Zhang_TARA_2023} propose TaRA, a model that addresses target inconsistency by constructing a Target Relation Graph and using mBERT with Graph Attention Networks to learn enriched, target-aware representations. Through contrastive learning, TaRA aligns semantically related targets across languages, enabling stance detection without direct training data. Results show strong performance on multilingual datasets, highlighting the importance of structured relational modelling and contrastive techniques for cross-lingual stance generalization.

\section{Challenges and Limitations}\label{Section:challenges} 

Despite growing research interest and practical applications, stance detection remains a challenging task with several technical, linguistic, and practical limitations.
One of the foremost challenges is data imbalance, where certain stance categories, such as ``comment'' or ``support," are significantly overrepresented, while critical categories such as ``deny'' and ``query'' are underrepresented. This class imbalance often leads to biased model predictions and poor generalization on underrepresented stances. Although data augmentation techniques such as synonym replacement, paraphrasing, or back-translation are employed to address this issue, they are typically general-purpose and fail to generate stance-specific variations, thus offering only marginal improvements.
The advent of Large Language Models (LLMs) such as GPT and Llama has revolutionised many natural language processing tasks, including stance detection. These models offer pre-trained, powerful representations of language. However, these models are not inherently fine-tuned for capturing the stance expression, especially in contexts where the stance is implicit, sarcastic, or heavily context-dependent. They often overlook stance-specific semantic cues or misinterpret rhetorical or emotionally charged language. Moreover, LLMs have high computational costs and memory requirements, limiting their accessibility for low-resource settings or real-time deployment.
A significant linguistic limitation comes from how people express their opinions indirectly, especially on platforms like Twitter and Reddit. Users frequently express their stance through irony, sarcasm, rhetorical questions, or cultural and political allusions, none of which are easy to capture using surface-level lexical or syntactic features. Detecting such indirect cues demands a deeper understanding of intent and social context, which remains beyond the scope of many current models.
Domain and topic sensitivity also limit the generalizability of stance detection systems.
Due to changes in language use, discourse patterns, or stance framing, models trained on one dataset or topic frequently perform poorly on another, such as vaccination. The practical scalability of these systems across real-world events and new themes is limited by the absence of efficient domain adaptation mechanisms.
Furthermore, one of the fundamental obstacles in stance detection is still explainability. While modern deep learning models perform well in terms of performance, they often act as black boxes, providing predictions without explaining the reasoning behind them. This lack of transparency is particularly problematic in sensitive areas such as misinformation detection, where stakeholders ranging from platform moderators to policymakers need clear justifications for model decisions.
Finally, resource limitations, especially in the creation of annotated datasets, pose a significant hurdle. Annotating stance in social media content is time-consuming and requires human expertise, making it difficult to scale for large, diverse datasets. Different annotators may have their own views when labelling stances, which can lead to inconsistencies and make it harder to build reliable models.

\section{Future Directions}\label{future}
The rapid advancements in stance detection, particularly with the integration of large language models (LLMs), open several promising research avenues. In this section, we outline key future directions that can further enhance the robustness, generalizability, and real-world applicability of stance detection systems.
\subsection{Enhancing Multimodal and Cross-Modal Understanding}
Future work should focus on improving multimodal stance detection by better aligning textual, visual, and auditory cues. While current models leverage images and text, they often struggle with interpreting sarcasm, memes, and contextual humour in multimedia content. Advanced vision-language models such as CLIP, Flamingo could be fine-tuned to capture subtle inter-modal dependencies, while graph-based approaches could model relationships between different modalities more effectively.
\subsection{Improving Zero-Shot and Few-Shot Generalization}
Despite progress in zero- and few-shot stance detection, models still face challenges when applied to unseen topics or low-resource languages. Future research could explore meta-learning, prompt-tuning, and synthetic data generation such as using LLMs like GPT-4 or Llama 3 to create diverse training examples. Additionally, contrastive learning and knowledge distillation techniques could help transfer stance-related knowledge across domains without extensive retraining.
\subsection{Cross-Cultural and Multilingual Adaptation}
Public opinion varies significantly across cultures, languages, and regions. Future research should investigate culturally aware stance detection models that account for linguistic nuances, regional slang, and sociopolitical contexts. Retrieval-augmented generation (RAG) or language-specific adapters could be added to multilingual LLMs like mBERT and XLM-R to enhance performance in low-resource languages.
\subsection{Addressing Bias and Fairness}
Biases inherited from training data might influence stance detection methods, particularly in politically or socially sensitive scenarios. Future studies should handle such scenarios by focusing on adversarial training, fairness-aware learning, and debiasing techniques. Furthermore it should ensure that models remain unbiased and equitable for better performance of stance detection models.
\subsection{Integration with Fact-Checking and Misinformation Detection}
By identifying the type of stance, stance detection can be very helpful in combating misinformation. Future systems may integrate stance identification with fact-checking pipelines, employing knowledge graphs and real-time evidence retrieval to verify claims and measure public opinion.

A notable emerging trend in stance detection is the shift toward {reasoning-centric and verifiable stance modeling}. Recent research increasingly treats stance detection not merely as a classification problem. The task is considered as a structured reasoning task that requires intermediate decision steps, explicit evidence grounding, and faithful explanation generation. This includes agentic frameworks that convert stance prediction into planning, retrieval, and reasoning phases, as well as verification-aware models.
These trends signal a transition toward adaptive, explainable, and reliable stance detection systems.

\section{Conclusion}
Stance detection has emerged as a crucial task in natural language processing (NLP), enabling the automated analysis of opinions, beliefs, and attitudes expressed in text, particularly in dynamic and opinion-rich environments such as social media. 
Large Language Models (LLMs) have revolutionized the area of stance detection by utilizing their cross-domain adaptability, contextual knowledge, and transfer learning.
The survey provided a comprehensive overview of stance detection, covering its evolution from traditional rule-based and statistical methods to modern LLM-driven approaches. We explored various learning paradigms such as supervised, unsupervised, zero-shot, and few-shot, highlighting their strengths and limitations in different scenarios. Additionally, we examined unimodal, multimodal, and hybrid approaches, emphasizing how integrating multiple data modalities such as text, images, and metadata, enhances stance detection performance.
A key focus was on target-based approaches such as in-target, cross-target, and multi-target stance detection demonstrating how models generalize across different topics and domains. We also reviewed benchmark datasets, evaluation metrics, results and real-world applications, including fake news detection, political discourse analysis, public health monitoring, and social media policy enforcement.
In conclusion, stance detection powered by LLMs holds immense potential for understanding public opinion, combating misinformation, and enhancing decision-making across various sectors. The continued evolution of stance detection promises to deepen our understanding of human communication in the digital age.
\bibliographystyle{elsarticle-num}

\bibliography{biblography}
\end{document}